\documentclass[11pt,a4paper]{article}
\usepackage[hyperref]{emnlp2020}
\usepackage{times,latexsym}
\usepackage{url}
\usepackage[T1]{fontenc}
\usepackage{microtype}

\usepackage{xspace,mfirstuc,tabulary}
\usepackage{times}
\usepackage{latexsym}
 \usepackage{tabu,array,multirow,graphicx}
 \usepackage{float}
\usepackage{url}
\usepackage{subcaption}
\usepackage{amsmath}
\usepackage{graphicx}
\usepackage{amsmath}
\usepackage{amssymb}
\usepackage{footmisc}
\usepackage{multirow}
\usepackage{caption}
\usepackage{subcaption}
\usepackage{booktabs}
\usepackage{todonotes}

\usepackage{algorithm}
\usepackage[noend]{algpseudocode}
\usepackage{mathtools}

\usepackage{booktabs}    
\makeatletter
\def\BState{\State\hskip-\ALG@thistlm}
\makeatother

\newcommand{\R}{{\mathbb R}}
\renewcommand{\L}{{\mathcal{L}}}

\renewcommand{\H}{{\boldsymbol H}}

\newcommand{\T}{{\boldsymbol T}}

\newcommand{\y}{{\boldsymbol y}}
\renewcommand{\u}{{\boldsymbol u}}
\renewcommand{\v}{{\boldsymbol v}}
\newcommand{\w}{{\boldsymbol w}}
\renewcommand{\a}{{\boldsymbol a}}

\title{CLAR: A Cross-Lingual Argument Regularizer for Semantic Role Labeling}

\author{Ishan Jindal\textsuperscript{a},  Yunyao Li\textsuperscript{a}, Siddhartha Brahma\textsuperscript{b}{\thanks{\textsuperscript{b} Work done while at IBM Research}}, and Huaiyu Zhu\textsuperscript{a} \\
\textsuperscript{a}IBM Research, Almaden Research Center, CA 95120 \\
\textsuperscript{b}Google Research, Berlin, Germany\\ 
  {\tt ishan.jindal@ibm.com, \{yunyaoli, huaiyu\}@us.ibm.com},\\ {\tt sidbrahma@gmail.com}}
  
\date{}

\aclfinalcopy 

\begin{document}
\maketitle

\begin{abstract}

Semantic role labeling (SRL) identifies predicate-argument structure(s) in a given sentence. Although different languages have different argument annotations, polyglot training, the idea of training one model on multiple languages, has previously been shown to outperform monolingual baselines, especially for low resource languages. In fact, even a simple combination of data has been shown to be effective with polyglot training by representing the distant vocabularies in a shared representation space. Meanwhile, despite the dissimilarity in argument annotations between languages, certain argument labels do share common semantic meaning across languages (e.g. \emph{adjuncts} have more or less similar semantic meaning across languages). To leverage such similarity in annotation space across languages, we propose a method called \emph{Cross-Lingual Argument Regularizer} (\emph{CLAR}). CLAR identifies such linguistic annotation similarity across languages and exploits this information to map the target language arguments using a transformation of the space on which source language arguments lie. By doing so, our experimental results show that CLAR consistently improves SRL performance on multiple languages over monolingual and polyglot baselines for low resource languages. 

\end{abstract}

\section{Introduction}

\begin{figure}
  \centering
  \includegraphics[width=\columnwidth]{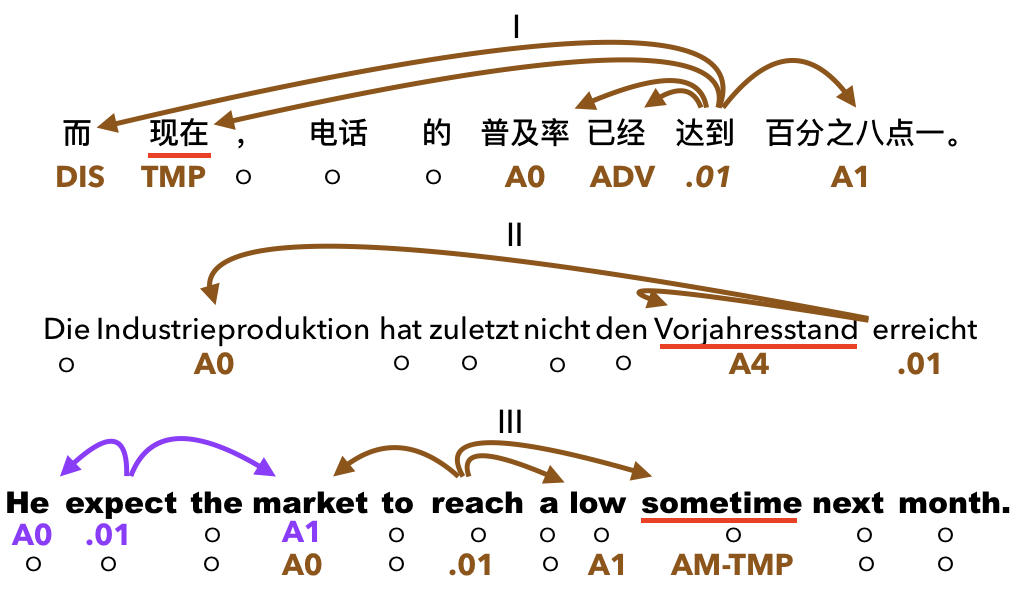}
  \caption{Example of predicate-argument structure from the CoNLL 2009 training data for I) Chinese, II) German, and III) English.}
  \label{fig:example}
\end{figure}%

Semantic Role Labeling (SRL) is the task of labeling each predicate and its corresponding arguments in a given sentence. SRL provides a more stable meaning representation across syntactically different sentences and has been seen to help a wide range of NLP applications such as question answering \cite{maqsud2014nerdle,yih2016value} and machine translation \cite{shi2016knowledge}.

Recent end-to-end deep neural networks for SRL, though performing well for languages with large training data \cite{marcheggiani2017simple,tan2018deep,he2018jointly}, are much less effective for low resources languages due to very limited annotated data for these languages. Methods such as \emph{polyglot training} \cite{mulcaire2018polyglot} seek to make these models perform better on low resource languages by combining supervision from multiple languages. The key idea in polyglot training is to combine the training data from multiple languages by using multilingual word embeddings from a shared space and a common encoder model (e.g. an LSTM). The argument sets for the languages are kept separate by using different classification layers. The arguments sets are kept separate because the semantic label spaces are usually language-specific ~\cite{mulcaire2018polyglot}. 

However, despite the dissimilarity in argument annotations between languages, certain argument labels do share common semantic meaning across languages. Fig. \ref{fig:example} shows three different sentences from Chinese, German, and English, respectively, with defined predicate-argument structures. Although the predicates are essentially the same, their arguments are labeled differently across languages in the training data. For instance, all sentences contain words representing the same underlying semantic meaning that is temporal but with different argument labels ({\small{\texttt{TMP}}} in Chinese, {\small{\texttt{A4}}} in German, {\small{\texttt{AM-TMP}}} in English).

We hypothesize that we can improve the SRL performance of low resource languages during cross-lingual transfer by identifying such arguments with similar semantic meaning across languages and representing them close to each other in the feature space. This requires: 
(1) Detecting the correspondence between the labels in different languages; and 
(2) Representing arguments with similar semantic meaning in the feature space for better SRL performance.

We  propose a method called \textit{Cross-Lingual Argument Regularizer} (CLAR) with a two-step process: \\

\noindent\textbf{Step 1: Pair Matching}: Detecting a number of label pairs between the source and target languages during polyglot training. 
We call these arguments \emph{common argument}s.  
Given the multilingual embedding already used in polyglot training, CLAR does not require additional cross-lingual alignments on parallel data.\\

\noindent\textbf{Step 2: Regularization}: Given the common arguments identified, find a transformation to bring the paired arguments close together. This transformation is learned and used in the polyglot training process so that the knowledge on the labels in the source language can be better transferred to knowledge in the corresponding labels in the target language.  

We evaluate CLAR on the SRL portion of the CoNLL 2009 dataset \cite{hajivc2009conll}{\footnote{We do not evaluate CLAR on Japanese data due to licensing issues.}} and compare its performance against baseline and polyglot training methods.

\begin{figure*}
  \centering
  \includegraphics[width=\textwidth]{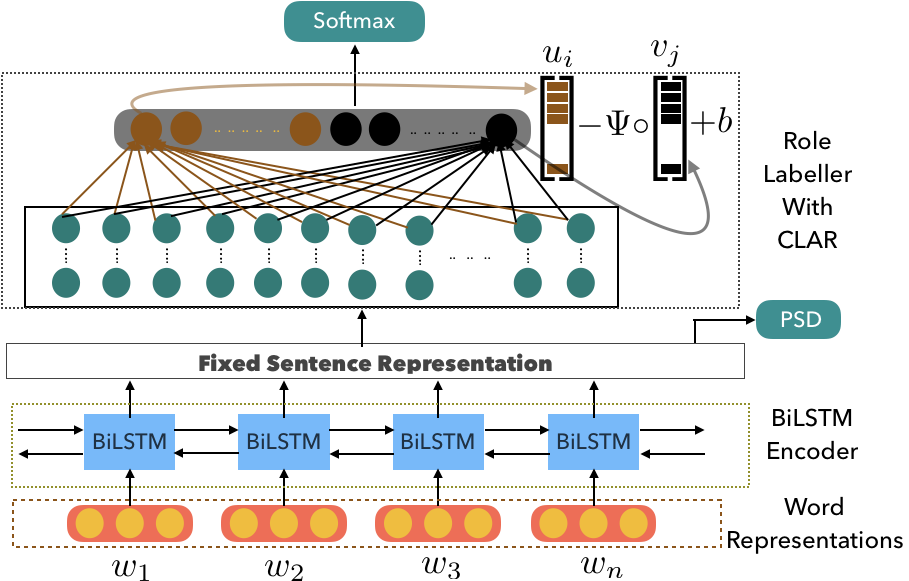}
  \caption{A multitask framework for predicate sense disambiguation and argument classification with CLAR argument regularization}
  \label{fig:model}
\end{figure*}
The main contributions of this work are:
\begin{itemize}
    \item We propose CLAR, a simple yet effective method for better cross-lingual transfer by detecting similar semantic role arguments between languages without requiring cross-lingual alignments or parallel data, and by learning a transformation for paired labels via regularization during SRL model training. 
    \item We conduct comprehensive empirical studies and demonstrate the effectiveness of CLAR over both monolingual and polyglot baselines.  
    \item We perform the ablation study and detailed analysis to understand why CLAR leads to better cross-lingual transfer and how its performance differs with different levels of correspondence among arguments.
\end{itemize}

The rest of the paper is organized as follows: Sec.~\ref{sec:basemodel} describes the base model. Sec.~\ref{sec:clar} describes CLAR.   Sec.~\ref{sec:exp} demonstrate its efficacy with extensive empirical evaluation.  Sec.~\ref{sec:review} reviews the existing literature.  Sec.~\ref{sec:conc} makes concluding remarks.


\section{Base Model}
\label{sec:basemodel}

The SRL task consists of four subtasks: 1) \textit{predicate identification} (e.g., \emph{reach}); 2) \textit{sense disambiguation} of the identified predicate (e.g., \emph{reach}.01); 3) \textit{argument identification} for each predicate (e.g., \emph{market}) and 4) \textit{role classification} of the identified arguments (e.g., {\small{\texttt{A0}}}).  Following \citet{li2018unified} and \citet{mulcaire2018polyglot}, we focus on argument labeling and predicate sense disambiguation, both sequence tagging problems. \\

\noindent\textbf{Model Architecture} As shown in Fig. \ref{fig:model}, our model architecture consists of four main modules: (1) \textit{sentence encoder} takes the raw tokens sequentially and outputs a fixed sentence representation; (2) \textit{role labeler} takes the sentence encoder output and identify and predicts roles of the tokens; (3) \textit{predicate sense disambiguator} takes the sentence encoder output and predict the sense for each predicate; and (4) \textit{CLAR regularizer} first detects the common arguments and then learns a manifold on which the arguments of the target languages lie. We now describe each of the modules in more details. 

\subsection{Sentence Encoder}
\textbf{Word Representation} Knowing the predicate position has previously been shown to improve the argument labeling task \cite{li2018unified} and since the predicate position is marked in the CoNLL 2009 dataset, we use this information and obtain the predicate-specific word representations for each word in the sentence. In addition to predicate-specific flag $\w_i^f$, we represent each word $\w_i$ in the sentence as a concatenation of several word features including randomly initialized word embeddings $\w_i^r$, pre-trained word embeddings $\w_i^p$, randomly initialized lemma embeddings $\w_i^l$ and randomly initialized POS tags embeddings $\w_i^s$. Finally, each word is represented as $\w_i = [\w_i^r,\w_i^p,\w_i^l,\w_i^s,\w_i^f]$. Since we combine the resources from a pair of languages similar to polyglot training \cite{mulcaire2018polyglot}, we use the language-specific pre-trained word embeddings for $\w_i^p$ and train the SRL model on the source and target language simultaneously.\\

\noindent\textbf{BiLSTM Encoder} To model the sequential input we use Bi-directional Long Short Term Memory neural networks \cite{hochreiter1997long}, which take in concatenated word representation for each word in the $j$-th sentence $x_j~=~(\w_{j1},~\w_{j2},~\cdots,~\w_{jn})$ and process them sequentially from both directions to obtain the contextual representations. 

\subsection{Semantic Role Labeler} Our role labeler consists of Multi-Layer Perceptron (MLP) layers with highway connections \cite{srivastava2015training}. It takes the contextualized word representations from the sentence encoder as an input and outputs a probability distribution over the set of argument labels for each word in the sentence. Given a sentence, we maximize the likelihood of labels for each word by minimizing
\begin{equation}
\mathcal{L_{\text{Base}}} = -\frac{1}{N}\sum_{i=1}^N p(y^\prime = y_i|\w_i;\theta),
\label{eq:crossentropy}
\end{equation}
where $y_i$ is the argument label, $\w_i$ represents the input token, $\theta$ represents the model parameters, and $N$ denotes the total number of samples.

\section{The CLAR Algorithm}
\label{sec:clar}

The underlying motivation for polyglot training \citet{mulcaire2018polyglot} is that arguments from different languages often help enhance each other. It is reasonable to assume that if corresponding arguments from source and target languages are located closer in the feature space, their mutual enhancements can be strengthened. The possibility for doing so is based on the following observation.

In neural network models that generate labels, the last layer is usually a softmax layer of the form
\begin{gather}
    \y_i = \frac{\exp({\H \a_i})}{\sum \exp({\H \a_i})}
\end{gather}
where $\y_i\in\R^k$, its $k$ components corresponding to the $k$ output argument labels. Given $\a_i\in\R^m$ as a representation of the input token $i$ calculated by previous layers, the rows $\boldsymbol{h}_k$ of the weights $\H$ are responsible for distinguishing the different argument labels~$k$ from each other. During the simple polyglot training, the $k$ argument labels consist of $k_s$ for the source language and $k_t$ for the target language.  Splitting these $\boldsymbol{h}_i$s into two sets, $\u_i$ for the source language and $\v_i$ for the target language, we observe that for arguments labels, the Euclidean distance between $\u_i$ and $\v_j$ are often small if the $i$ and $j$ are corresponding argument labels. These can be brought even closer together by an affine transform (linear transform and translation).  

We therefore propose the following approach (CLAR) consisting of two steps:
\begin{description}
\item[Step 1: Pair Matching:] Detect the best pairing of the arguments between a pair of languages. 
\item[Step 2: Regularization:] Find a transformation that brings the feature vectors corresponding to the paired argument labels close to each other.
\end{description}
These two steps are described in detail below.\\

\noindent\textbf{Pair Matching:} The goal of this step is to identify matching label pairs in the two languages. We start with the simple polyglot training \cite{mulcaire2018polyglot} for the first few epochs without CLAR and collect the last layer weights for all the target and source language arguments.

Given the $k_s$ vectors $\u_i$ and $k_t$ vectors $\v_j$, solve this constraint optimization problem
\begin{equation}
\begin{aligned}
 \underset{\T}{\text{minimize}}
 &  \sum_i^{k_s}\sum_j^{k_t} \T_{ij}||\u_i - \v_j||_2^2 \\
 \text{subject to}\\
 & \sum_i \T_{ij} \leq 1,  \; j = 1, \ldots, k_t \\
  & \sum_j \T_{ij} \leq 1,  \; i = 1, \ldots, k_s \\
   & \sum_{i,j} \T_{ij} \geq \text{min}(k_t, k_s),  \;  j = 1, \ldots, k_t;\\
   & \qquad \qquad \qquad \qquad \qquad i = 1, \ldots, k_s\\
    & \T_{ij} \in \{0,1\},  \; \forall i,j.
\end{aligned}
\label{eq:objective}
\end{equation}
Intuitively, this requires finding pairings between $i$ and $j$ such that the total squared distance between paired vectors $(\u_i, \v_j)$ is minimized, subject to the constraint that each source argument matches at most one target argument and vice versa, and that at least $K = \text{min}(k_t, k_s)$  argument pairs are identified. 
This identifies $K$ semantically similar argument pairs in source and target languages, represented in the binary matrix $\T$, where $\T_{ij} = 1$ means that argument $i$ in source language and argument $j$ in target language are paired together. 
Later on (Sec.~\ref{subsec:analysis}) we will show that in certain situations it makes sense to relax the ``at most one'' constraint and allow many-to-one or one-to-many matching.

This is an \textit{Integer Linear Programming} problem, for which many excellent solvers exist. We use GLPK solver from CVXOPT{\footnote{http://cvxopt.org/index.html}}.

We observe that the frequency distribution of the argument labels is quite skewed in the training dataset: a few labels (e.g., {\small{\texttt{A0}}}, {\small{\texttt{A1}}}) have much larger number of training examples than other labels. Experiments show that low-frequency labels cause noisy pair matching that degrades the output quality.  Therefore, we consider only labels that have more than $1\%$ of the total number occurrences in the respective language training data. Typically, $40-50{\%}$ of the total labels in each language match this criterion.  The $k_s$ and $k_t$ in the general algorithm are replaced by $\hat k_s$ and $\hat k_t$ for the number of arguments satisfying this criterion in the source and the target language, respectively.\\

\noindent\textbf{Regularization:}
The goal of this step is to learn an affine transform to bring the target vectors closest to the corresponding source vectors.  This step is performed iteratively during the overall training process.

Given the $K$ pairs  $(\u_i, \v_i)$ detected in the previous step, the objective of the overall optimization objective function is amended as follows
\begin{equation}
    \L_{\text{CLAR}} = \L_{\text{Base}} + \lambda\sum_{i=1}^{K}||\u_i - \Psi \v_i+b||_2^2,
\label{eq:final}
\end{equation}
where $\Psi \v_i+b$ is the affine transform to bring $\v_i$ close to $\u_i$,  and $\lambda$ controls the strength of the amendments by the paired labels.
The transformation $\Psi, b$ is learned iteratively by minimizing \eqref{eq:final} during SRL model training.

\section{Experiments}
\label{sec:exp}
\begin{table*}[h!]
\footnotesize
\begin{center}
\begin{tabu}{llccccccc}
\toprule
 & &EN&CA&CS&DE&ES&ZH&avg \\  \midrule
 \multicolumn{2}{l}{\citet{zhao2009multilingual}}& 86.20& \underline{\textbf{80.30}} &  85.20 & 76.00& 83.00& 77.70&  - \\
 \multicolumn{2}{l}{\citet{roth2016neural}}&87.70& - &  - & \underline{\textbf{80.10}}&  80.20& 79.40&  - \\
\multicolumn{2}{l}{\citet{marcheggiani2017simple}}&87.70& - &  \underline{\textbf{86.00}} & -&  80.30& 81.20&  - \\
\multicolumn{2}{l}{\citet{cai2018full}}& 89.60&- &  - & -&  -& \underline{\textbf{84.30}}&  - \\
\multicolumn{2}{l}{\citet{kasai2019syntax}}& \underline{\textbf{90.20}}&- & - & -&  \underline{\textbf{83.00}}&-&  - \\ 
\multirow{2}{*}{\citet{mulcaire2018polyglot}}& Monolingual&86.54&77.31&84.87&66.71&75.98&81.26&  77.22\\
& Polyglot&-&79.08&84.82&69.97&76.45&81.50&  \underline{\textbf{78.36}}\\ 
\toprule
\multirow{3}{*}{Base SRL + MUSE Embedding}& Monolingual&86.47&78.92&\textbf{89.78}&68.73&78.09&81.34& 79.37\\
& Polyglot&-&79.05&89.70&71.16&78.22&81.42&  79.78\\
& CLAR&-&\textbf{79.26}&89.77&\textbf{72.50}&\textbf{78.83}&\textbf{81.85}& \textbf{80.44 }\\
\toprule
\multirow{3}{*}{Base SRL + BERT Embedding}& Monolingual&88.14&80.50&90.78&74.39&80.98&84.71&  82.27\\
& Polyglot&-&81.87&90.67 &74.45&81.88&84.79&  82.73\\
& CLAR&-&\underline{\textbf{82.18}}&\underline{\textbf{90.81}} &\textbf{75.33}&\textbf{82.13}&\underline{\textbf{85.04}}&  \underline{\textbf{83.09}}\\
\bottomrule
\end{tabu}
\end{center}
\caption{Semantic F1 scores (including sense) on CoNLL 2009 Shared task languages. The best reported performance on English and Spanish from \cite{kasai2019syntax}, Chinese from \cite{cai2018full}, German from  \cite{roth2016neural}, Catalan from  \cite{zhao2009multilingual} and Czech from\cite{marcheggiani2017simple}. Underline shows the best performance among all methods.}
\label{table:clarpsdperformance}
\end{table*}

\subsection{Dataset}
We evaluate CLAR on CoNLL 2009 Shared Task dataset \cite{hajivc2009conll} with English (EN) as the source language and five different languages, namely German (DE), Spanish (ES), Chinese (ZH), Czech (CS) and Catalan (CA), as target languages. 

The dataset includes no correspondence defined between the argument labels across languages. For instance, the argument label set in English contains ({\small{\texttt{A0}}}, {\small{\texttt{A1}}}, $\cdots$) while the argument label set in Spanish contains (~{\small{\texttt{Arg0-agt}}},~{\small{\texttt{Arg0-pat}}},~$\cdots$). Further details on dataset is available in Appendix \ref{ap:datadesc}.

\begin{table}[h!]
\begin{center}
\begin{tabu}{lccccc}
\toprule
EN&+CA&+CS&+DE&+ES&+ZH \\ \midrule
86.47& 87.12&86.70&87.09&86.68&86.90 \\ \bottomrule
\end{tabu}
\end{center}
\caption{CLAR Semantic F1 scores (including sense) on EN test set for each language pair.}
\label{table:eng}
\end{table}

\begin{table*}[h!]
\footnotesize
\begin{center}
\begin{tabu}{lcccccccccc}
\toprule
 \multicolumn{1}{l}{\multirow{2}{*}{Target}} & \multicolumn{1}{c}{\multirow{2}{*}{Source}}&  \multicolumn{3}{c}{Monolingual}&  \multicolumn{3}{c}{Polyglot}&  \multicolumn{3}{c}{CLAR}\\ \cmidrule(lr){3-5} \cmidrule(lr){6-8} \cmidrule(lr){9-11}
  & &\textbf{P} & \textbf{R} & \textbf{F1}& \textbf{P} & \textbf{R} & \textbf{F1}& \textbf{P} & \textbf{R} & \textbf{F1}\\ 

\cmidrule(lr){3-5} \cmidrule(lr){6-8} \cmidrule(lr){9-11}
CA  &+EN& {78.47}  &  75.44  &  76.92&  77.59  &  {76.68}  &  77.13&78.35  &  76.54  &  \textbf{77.44}\\ 
CS  &+EN& {80.36} &  76.00  & 78.12&  80.32  &  75.69  &  77.93&79.91 &  {76.50}   &  \textbf{78.17}\\ 
DE  &+EN&69.64     &  64.43     &  66.94&  71.66  &  69.96  &  70.80& {73.10}  &  {71.54}  &  \textbf{72.31}\\ 
ES  &+EN& 78.22  &  75.63  &  76.90&  78.37  &  75.83  &  77.07&{79.77}  &  {76.22}   &  \textbf{77.95}\\ 
ZH  &+EN& 78.27  &  75.07  &  76.64&  79.04  &  74.50  &  76.68&{79.36}  &  {75.43}  &  \textbf{77.34}\\
\bottomrule
\end{tabu}
\end{center}
\caption{CLAR performance (argument classification only) on CoNLL 2009 Shared task languages and comparison with polyglot and monolingual methods.}
\label{table:clarperformance}
\end{table*}

\subsection{Setup}
We compare CLAR with several \emph{Monolingual} and \emph{Polyglot} methods. For monolingual baselines, we train separate SRL models for each language.
For Polyglot and CLAR methods, we train the SRL model on a pair of language. We use pre-trained multilingual embeddings to allow the multilingual sharing between languages. We use Multilingual Unsupervised and Supervised Embeddings (MUSE) \cite{conneau2017word} for all the languages except Chinese{\footnote {MUSE does not provide aligned vectors for the Chinese language}}, where we use fastText aligned word embeddings \cite{joulin2018loss}. We also use the pre-trained BERT multilingual cased embeddings \cite{devlin2019bert} in place of MUSE pre-trained embeddings to observe the effect of better multilingual embeddings. Details on model hyperparameters are presented in Appendix \ref{ap:hyper}. 
For all the experiments we fix the base model architecture. For the Polyglot training, we implement the \emph{simple polyglot sharing} setup proposed by \citet{mulcaire2018polyglot}. Along with the reported results in \citet{mulcaire2018polyglot} we also report the polyglot results with our model architecture keeping the same word representation to avoid any ambiguity between Polyglot and CLAR comparison.

\subsection{Results}
\noindent\textbf{Comparison Against Polyglot and Monolingual Training:} Table \ref{table:clarpsdperformance} summarizes the performance of CLAR and all baselines for SRL. As can be seen, for both MUSE and BERT embeddings, CLAR results in better SRL models than those obtained via monolingual and polyglottraining for all target languages. The improvement is particularly noticeable for the languages with much fewer (< 1/3) training samples than those of EN (e.g. DE and ES). This result confirms that CLAR can effectively transfers knowledge from a high resource language (EN) to other languages with less resource.

Note that for CS, neither CLAR nor polyglot training shows performance gain over the baseline. CLAR outperforms the polyglot baseline but remains on par with the monolingual baseline. 
We present further investigation on this in Section \ref{subsec:analysis}.  \\

\noindent\textbf{Comparison Against SoTA:} With the powerful BERT multilingual embeddings, CLAR surpasses the best previously reported results on 3 out of 6 languages (Table~\ref{table:clarpsdperformance}). In fact, its average performance surpasses that any previous-reported single system. The strong performance of CLAR confirms its great promise for cross-lingual transfer. \\

\noindent\textbf{Cross-Lingual Transfer from Target to Source Language:} Interestingly, cross-lingual transfer by CLAR also helps improving the performance of languages with abundant training data. As illustrated in Table \ref{table:eng}, transferring knowledge using CLAR from other languages to EN leads to small but consistent improvements for EN. \\

\noindent\textbf{CLAR Performance on Arguments Alone:} 
Since CLAR mainly affects role labeling, we conduct further analysis of its performance on argument classification alone (i.e. predicate sense disambiguation is not evaluated). The results are summarized in Table \ref{table:clarperformance} for Base SRL + MUSE embedding. 
One can observe that for all target languages, CLAR registers small but noticeable improvements ($0.24\%$ to $1.51\%$) for argument classification in comparison to both monolingual and polyglot methods. The consistent improvements confirm the effectiveness of CLAR in enabling better cross-lingual transfer. 

\begin{figure*}[h!]
\begin{subfigure}{.33\textwidth}
  \centering
  \includegraphics[width=\linewidth]{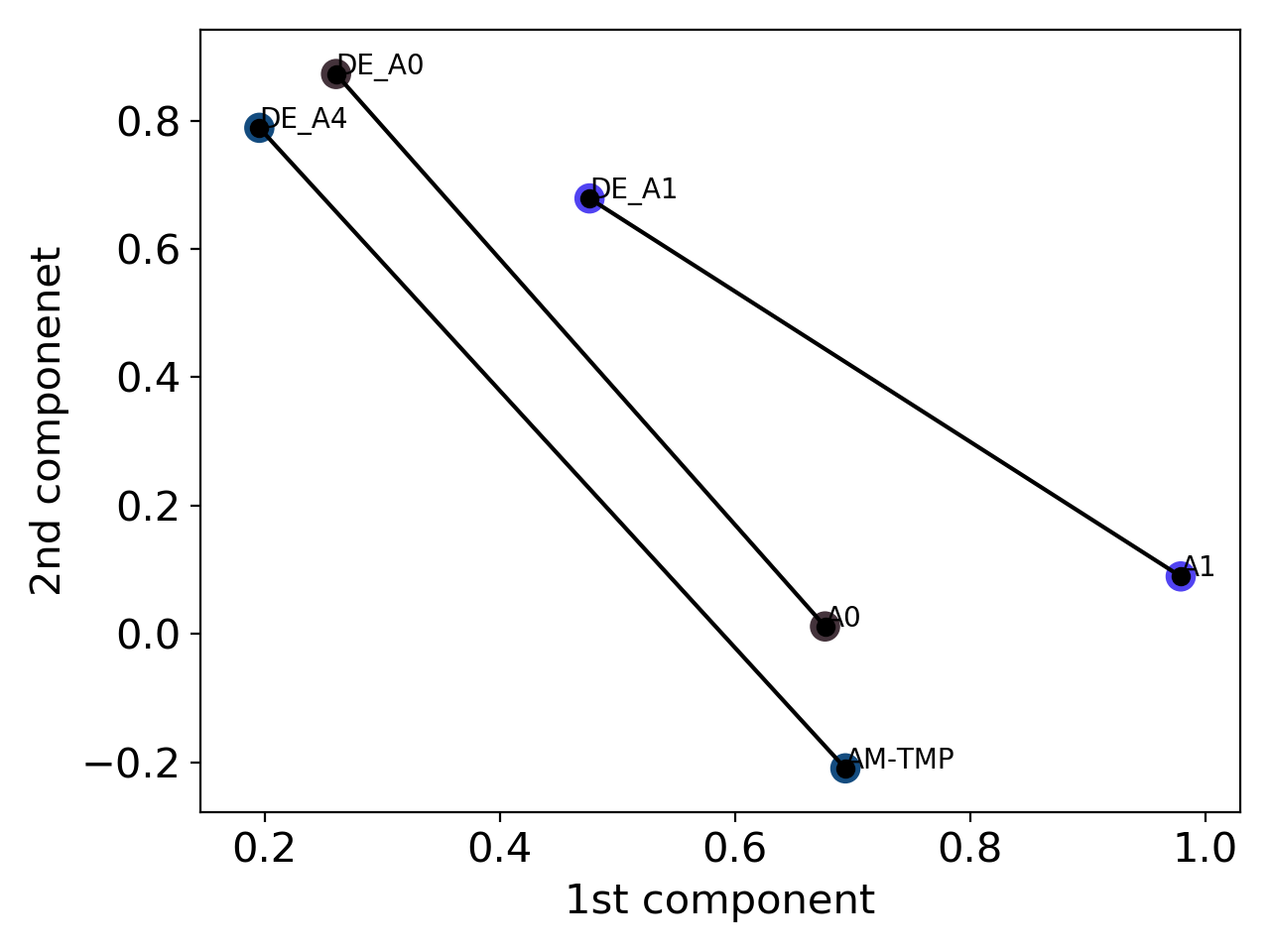}
  \caption{Polyglot German}
  \label{fig::polyDE}
\end{subfigure}%
\begin{subfigure}{.33\textwidth}
  \centering
  \includegraphics[width=\linewidth]{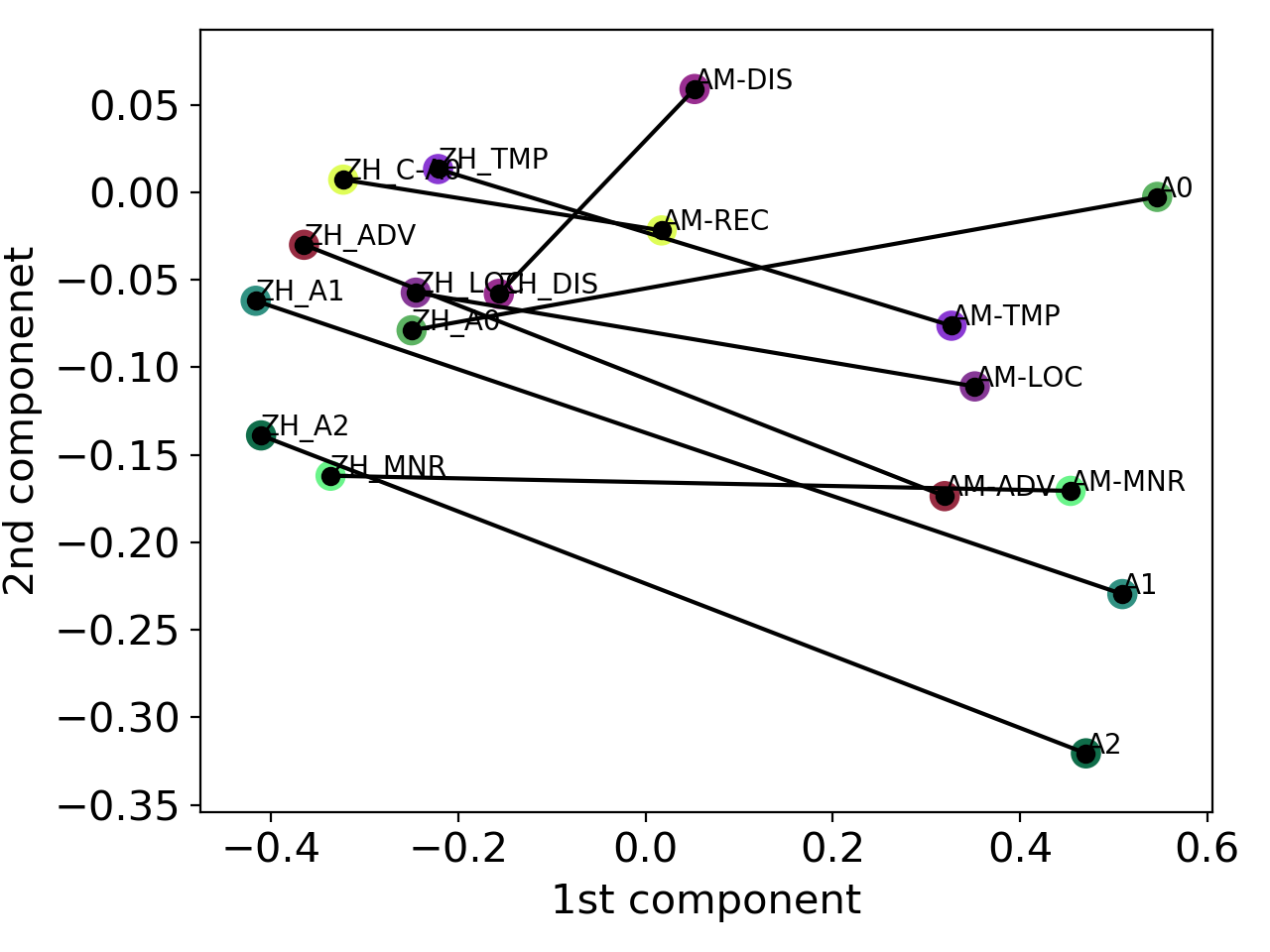}
  \caption{Polyglot Chinese}
  \label{fig::polyzh}
\end{subfigure}%
\begin{subfigure}{.33\textwidth}
  \centering
  \includegraphics[width=\linewidth]{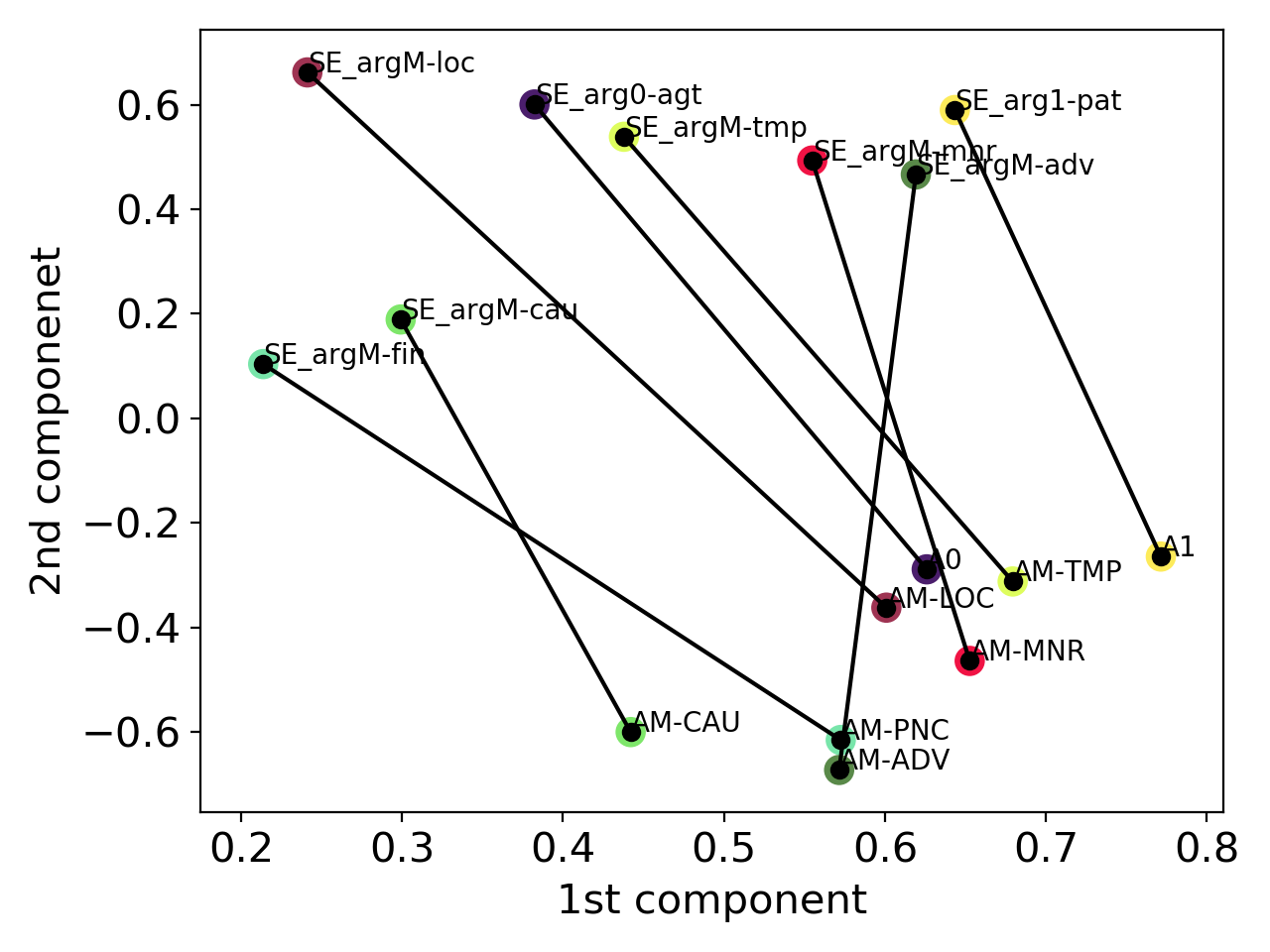}
  \caption{Polyglot Spanish}
  \label{fig::ployES}
\end{subfigure}
~
\begin{subfigure}{.33\textwidth}
  \centering
  \includegraphics[width=\linewidth]{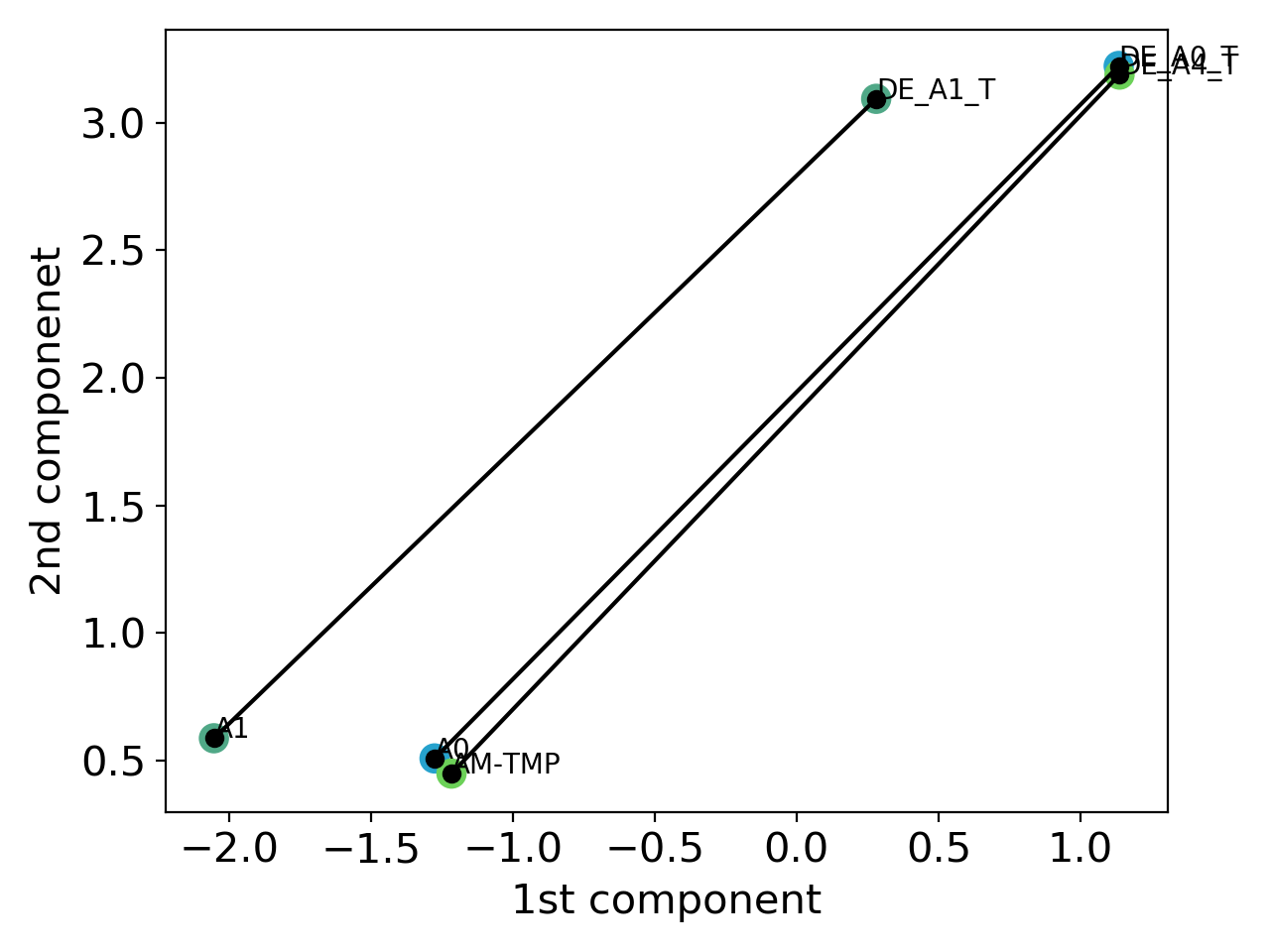}
  \caption{CLAR German}
  \label{fig::clarDE}
\end{subfigure}
\begin{subfigure}{.33\textwidth}
  \centering
  \includegraphics[width=\linewidth]{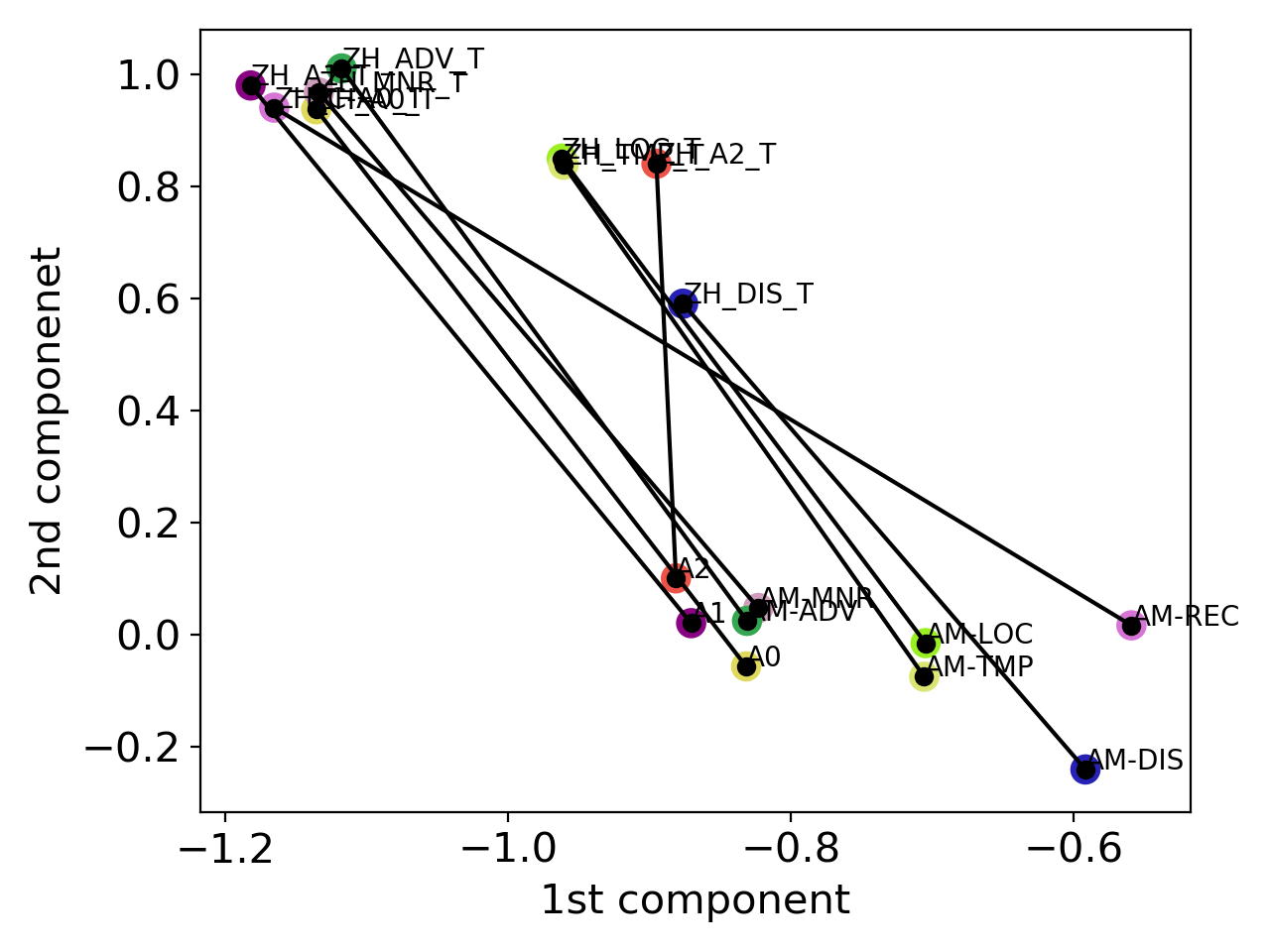}
  \caption{CLAR Chinese}
  \label{fig::clarZH}
\end{subfigure}%
\begin{subfigure}{.33\textwidth}
  \centering
  \includegraphics[width=\linewidth]{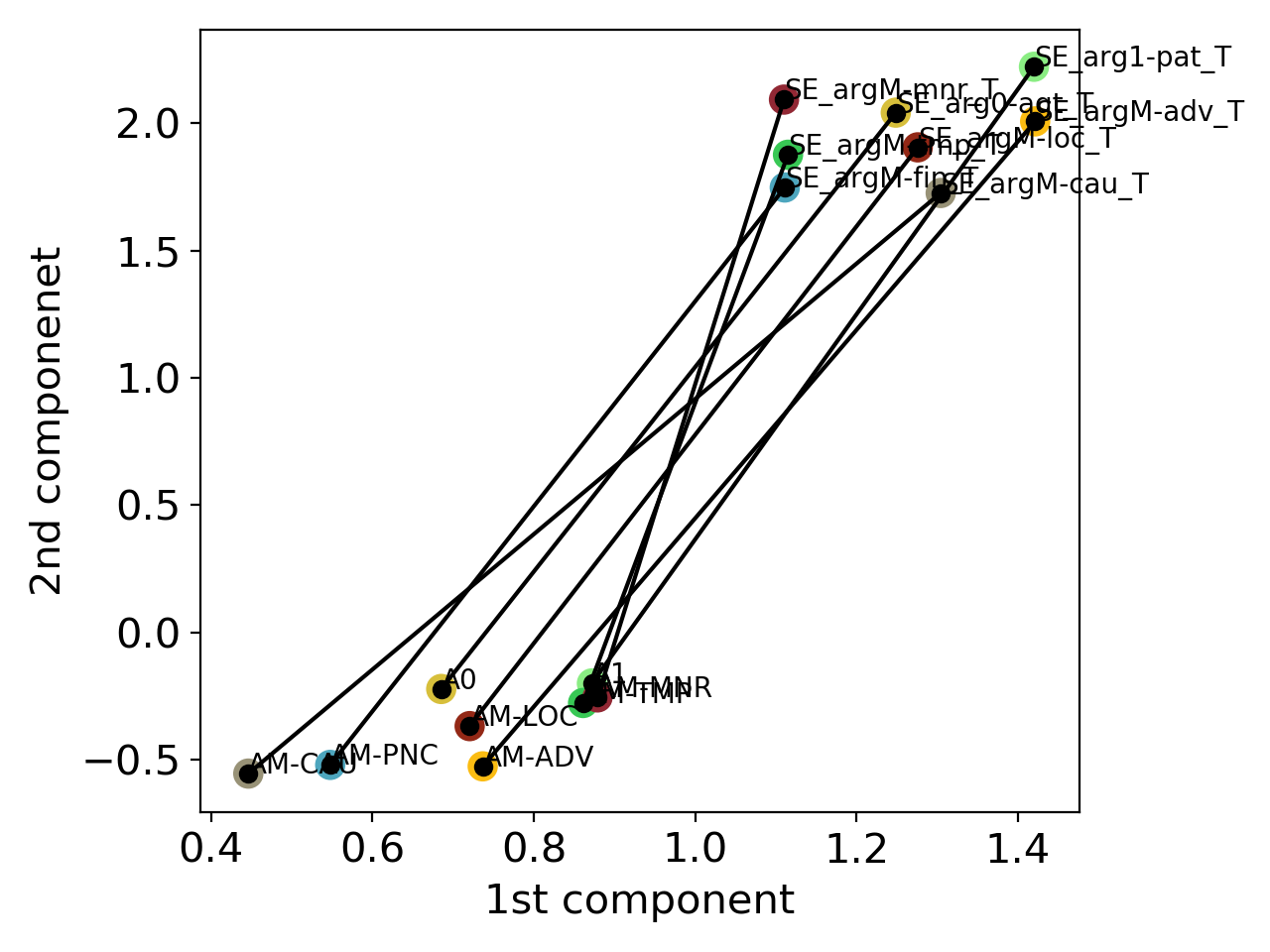}
  \caption{CLAR Spanish}
  \label{fig::clarES}
\end{subfigure}
\caption{A low dimensional representation of output layer weights of the matched arguments in source and target language as determined by polyglot learned weight vectors in row I and by \eqref{eq:objective} in row II.}
\label{fig:pca_clarpoly}
\end{figure*}

\subsection{What does CLAR do?}
\label{subsec:clar}
The results of our comparison studies clearly demonstrate that CLAR outperforms both baseline and polyglot training methods. In this subsection we first explain the intuition behind CLAR and then investigate how it regularizes the arguments.\\

\noindent\textbf{Intuition:} During Polyglot training we examine the last layer weights of the base SRL model and hypothesize that there exists a mapping between source and target language argument. To evaluate this hypothesis, we plot the weights of the output layer using SVD by keeping the two directions corresponding to top two largest eigenvalues learned by Polyglot (Row I) training in Fig. \ref{fig:pca_clarpoly}.

We draw a line between the arguments that are paired by Equation~\eqref{eq:objective}.
As can be seen, the euclidean distance between some of the paired arguments is similar. For instance, the euclidean distance between the arguments {\small{\texttt{A1}}} and {\small{\texttt{ZH-A1}}} is similar to that between {\small{\texttt{A2}}} and {\small{\texttt{ZH-A2}}} in Fig. \ref{fig::polyzh}. This pattern emerges from the training data for most of the target languages.

Further, we observe that the euclidean distances among the common arguments for the source and target languages are also similar. For example, in Fig. \ref{fig::polyzh},
the euclidean distance between the source (EN) arguments {\small{\texttt{A1}}} and {\small{\texttt{A2}}} is similar to that between the target language arguments {\small{\texttt{ZH-A1}}} and {\small{\texttt{ZH-A2}}}.
This observation holds true for most of the arguments across the target languages (Fig. \ref{fig::polyDE} - \ref{fig::ployES}). 

The above observations confirm that there exists similar arguments in source and target languages. The arguments in target language lie on a manifold that is similar in structure, with some translation and/or rotation, to the manifold on which the source language argument lies. \\

\noindent\textbf{Argument Matching and Regularization:}
Therefore, we first match the arguments with similar meanings in the target and the source language. We observe that almost all the matched argument pairs have similar meaning: some are syntactically visible (e.g. {\small{\texttt{ES-argM-adv}}} in ES and {\small{\texttt{AM-ADV}}} in EN), whereas others are semantically similar (e.g. {\small{\texttt{ES-argM-fin}}} and {\small{\texttt{AM-PNC}}} having the same meaning {\small{\texttt{purpose}}}). After obtaining the matched argument pairs, we regularize the output layer weights of the matched target arguments by forcing them to live on a matched source arguments manifold in \eqref{eq:final}. A list of matched arguments for various language pairs is provided in Appendix \ref{ap:pairedargs}.

We plot the CLAR learned weight vectors in Fig. \ref{fig:pca_clarpoly} (Row II). We can observe the uniformity in lines (in terms of length), which are drawn between paired target to source language arguments. Further, to quantify the length of these lines, we plot the euclidean distance matrix among the matched source language arguments. Among the target language arguments, we compute the correlation coefficient between the euclidean distance for EN-DE, EN-ZH, and EN-ES to be $0.9984$, $0.9531$, and $0.9352$ respectively. The fact that all these coefficients are close to one indicates that CLAR is indeed able to detect a manifold for the target language arguments similar to the one for the source language arguments. 
Our experimental results (Table~\ref{table:clarperformance}) demonstrate that allowing the paired target language arguments to lie on the detected manifold improves the argument classification performance. 

\subsection{Ablation and Analysis}
\label{subsec:analysis}
\textbf{Effect of $K$}  We also observe the impact of $K$ on the argument classification performance in Table \ref{table:Keffect}. We find that regularizing all the arguments obtained from \eqref{eq:objective}, while performing better than polyglot, is not a great choice overall. We suspect that considering all the paired arguments adds noise in the system. This is likely because some of the arguments in the target languages are language-specific and might be matched with an argument in the source language which has no close correspondence, for example, the Chinese argument {\small{\texttt{ZH-C-C-A0}}} has no direct corresponding argument in English. 

\begin{table}[t!]
\footnotesize
\begin{center}
\begin{tabu}{lcccc}
\toprule
Target&0&2&K/2&K \\ \midrule
CA&77.13&77.13&\textbf{77.44}&77.20 \\ 
CS&77.93&78.12&\textbf{78.17}&77.45 \\ 
DE&70.80&72.08&\textbf{72.31}&71.20 \\ 
ES&77.07&77.23&\textbf{77.95}&77.12 \\ 
ZH&76.68&76.87&\textbf{77.34}&77.02 \\ 
\bottomrule
\end{tabu}
\end{center}
\caption{Effect of $K$ on argument classification performance ($K=0$ represents Polyglot training)}
\label{table:Keffect}
\end{table}
Additionally, in some languages, arguments are labeled at a very granular level, and multiple arguments in these languages may correspond to a single argument in the source language. \\
For example, multiple arguments in Czech frequently map to only one corresponding argument in English.\\

\noindent\textbf{Languages with Similar Linguistic Annotations:} 
To further study the effectiveness of CLAR, we analyze the cross-lingual transfer between the languages known to have similar linguistic annotations. We expect to observe better cross-lingual transfer between such language pairs. Specifically, we examine Spanish (ES) and Catalan (CA) from the same AnCora corpus \cite{taule2008ancora}. 
We consider ES as the source language because it has more training samples than CA.

\begin{table}[t!]
\footnotesize
\begin{center}
\begin{tabu}{llccc}
\toprule
 \multicolumn{1}{l}{\multirow{1}{*}{Training}} & \multicolumn{1}{l}{\multirow{1}{*}{Method}}& \textbf{P} & \textbf{R} & \textbf{F1}\\ \midrule
CA  	&Baseline&  78.47  &  75.44  &  76.92\\  \midrule
\multirow{2}{*}{+ES} &Polyglot &  79.10  &  75.90  &  77.47\\
 &CLAR &  78.72  &  77.91  &  \textbf{78.31}\\
\multirow{2}{*}{+EN}  &Polyglot &  77.59  &  76.68  &  77.13\\
&CLAR &78.35  &  76.54  &  77.44 \\\bottomrule
\end{tabu}
\end{center}
\caption{Catalan argument classification performance with Spanish as source language}
\label{table:enca_clar}
\end{table}
In Table \ref{table:ESCA_args} we show the paired arguments detected by CLAR along with the euclidean distance between them. It can be seen that the euclidean distance for all paired arguments are close to 1, confirming that CLAR can effectively match semantically similar arguments across languages. 

The experimental results are summarized in Table \ref{table:enca_clar}. As expected, CLAR surpasses all prior results on CA. With the semantically similar language ES, the SRL performance on CA is better than the monolingual and polyglot training methods. Further, we observe a $0.87$ point absolute gain in F1 score when the cross-lingual transfer occurred from a similar linguistic annotated language (ES) than a less similar language (EN), despite of much smaller training data size ($\leq 30\%$ of EN). This observation strengthen our hypothesis that by representing the semantically similar arguments across languages on similar manifolds improves the SRL performance.
\begin{table}[t!]
\footnotesize
\begin{center}
\begin{tabu}{llr}
\toprule
Target &	Source &	Pair distance\\ \midrule
{\small{\texttt{CA-argM-tmp}}} &	{\small{\texttt{ES-argM-tmp}}} &	0.9302\\ 
{\small{\texttt{CA-argM-cau}}} &	{\small{\texttt{ES-argM-cau}}} &	0.9523\\ 
{\small{\texttt{CA-argM-atr}}} &	{\small{\texttt{ES-argM-atr}}} &	0.9542\\ 
{\small{\texttt{CA-arg2-ben}}} &	{\small{\texttt{ES-arg2-ben}}} &	0.9608\\ 
{\small{\texttt{CA-argM-fin}}} &	{\small{\texttt{ES-argM-fin}}} &	0.9657\\ 
{\small{\texttt{CA-arg1-null}}}&	{\small{\texttt{ES-arg1-null}}}&	0.9672\\ 
{\small{\texttt{CA-argM-mnr}}} &	{\small{\texttt{ES-argM-mnr}}} &	0.9709\\ 
{\small{\texttt{CA-argM-loc}}} &	{\small{\texttt{ES-argM-loc}}} &	0.9790\\ 
{\small{\texttt{CA-argM-adv}}} &	{\small{\texttt{ES-argM-adv}}} &	0.9810\\ 
{\small{\texttt{CA-arg0-cau}}} &	{\small{\texttt{ES-arg0-cau}}} &	0.9839\\ 
\midrule
\end{tabu}
\end{center}
\caption{Paired arguments in the source (ES) and the target language (CA)}
\label{table:ESCA_args}
\end{table}

\begin{figure}[t!]
  \begin{minipage}{0.5\columnwidth}
	\centering
	\includegraphics[width=1.1\linewidth]{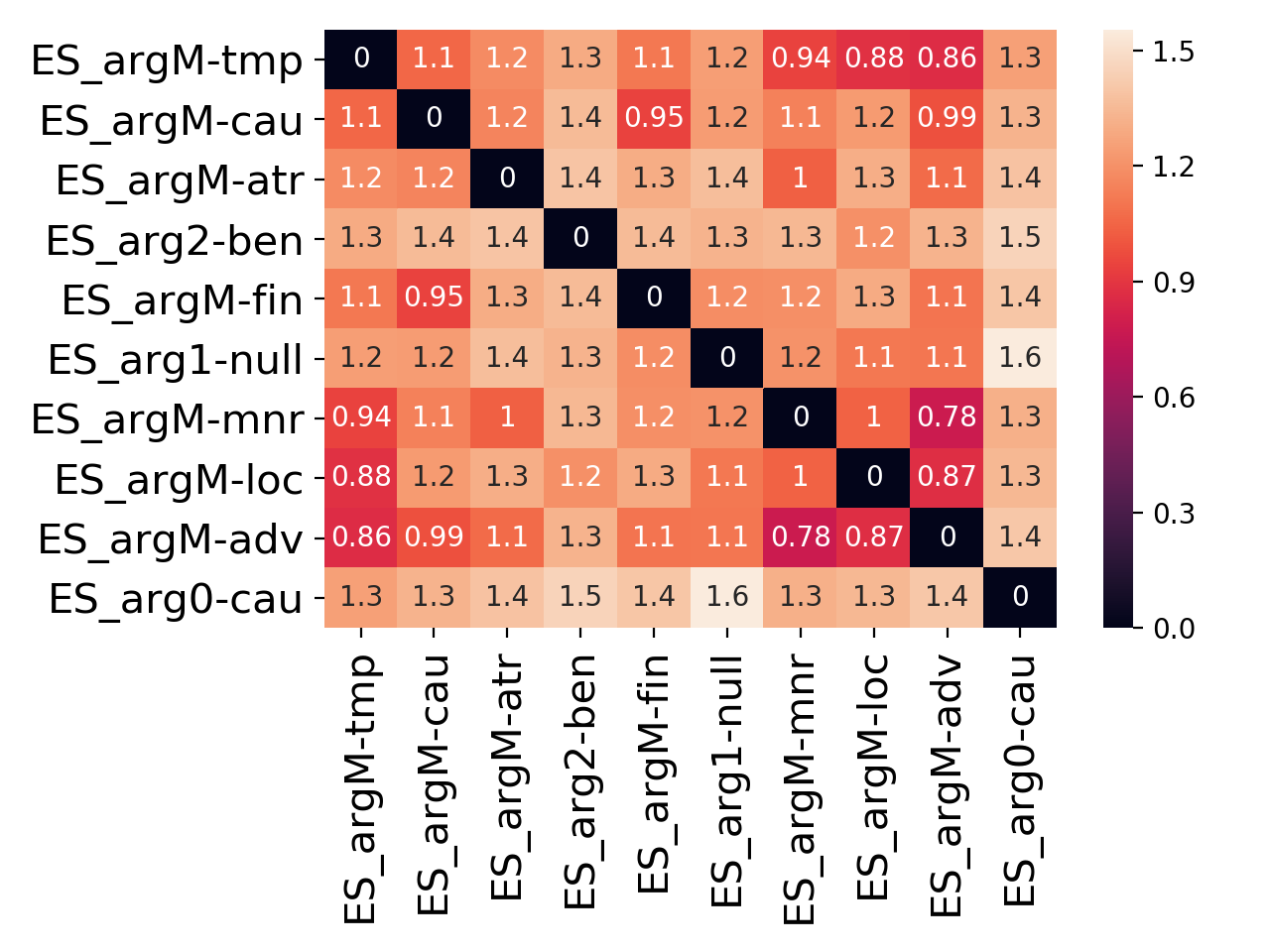}
	\subcaption{ES}
	\label{fig::esca_dist_es}
\end{minipage}%
  \begin{minipage}{0.5\columnwidth}
	\centering
	\includegraphics[width=1.1\linewidth]{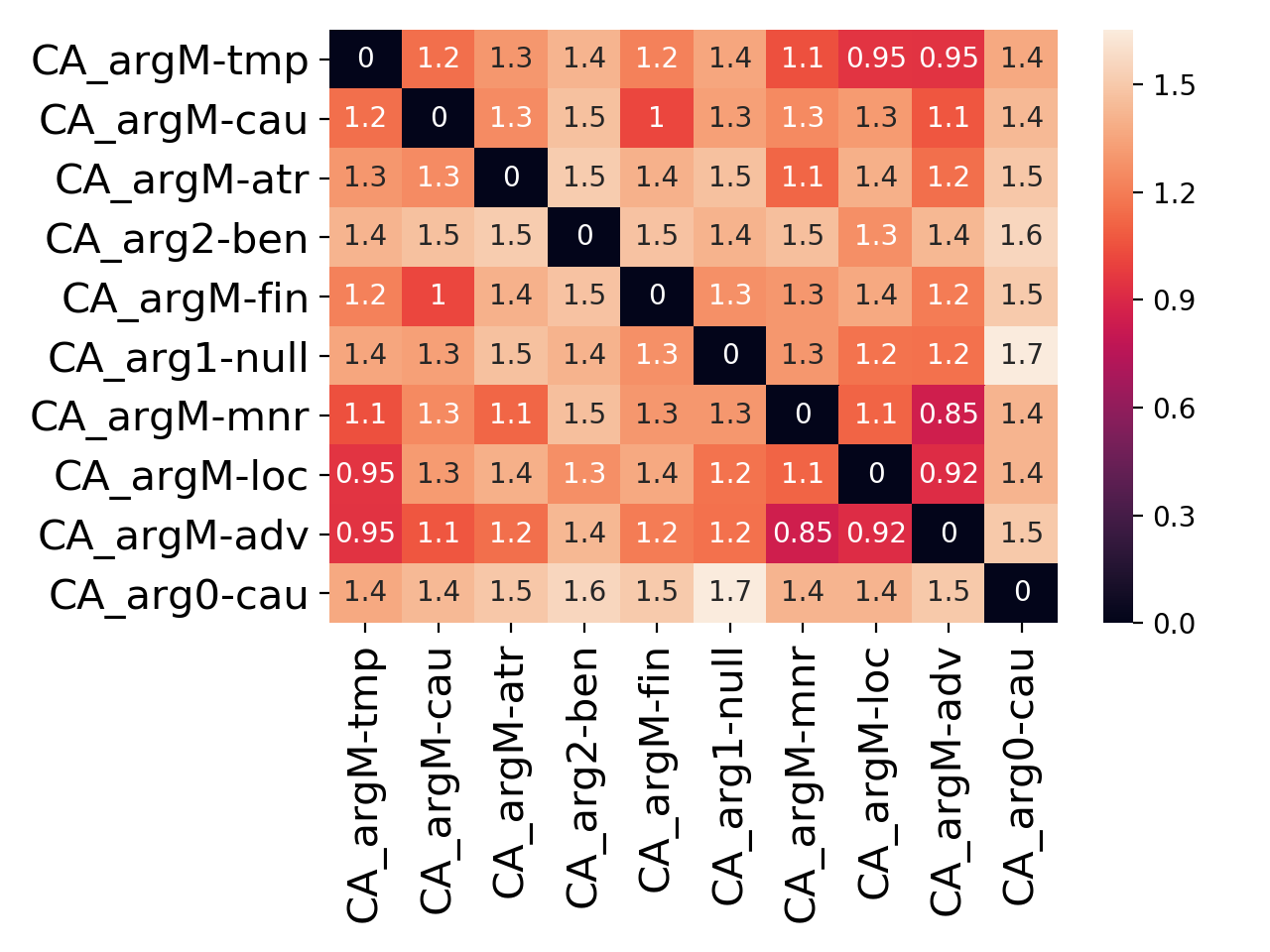}
	\subcaption{CA}
	\label{fig::esca_dist_ca}
\end{minipage}

\caption{Euclidean distance between last layer weights for ES-CA cross-lingual transfer.}
\label{fig:esca_dist}
\end{figure}

To visualize the space on which the common source and target language argument lies, we plot the heatmap of the euclidean distance between the last layer weights of the learned model in Fig. \ref{fig:esca_dist}. We plot the separate heatmaps among the paired arguments for each language, the source language (in Fig. \ref{fig::esca_dist_es}) and the target language (in Fig. \ref{fig::esca_dist_ca}). We observe these two heatmaps look very identical in distribution (a very high correlation coefficient $0.9996$ and a low Frobenius norm square of the difference $1.793$). This means that CLAR transforms the weight vectors of the corresponding target language arguments in such a way that the transformed weight vectors lie on a manifold, which is similar to another manifold on which source language argument weights lie but translated and/or rotated. The aforementioned is evident from Table \ref{table:ESCA_args} where we report the distance between these argument pairs.\\

\noindent\textbf{Why is Czech an Exception?}
Though Czech (CS) has the most training samples in the CoNLL 2009 dataset, the cross-lingual transfer to and from CS is not very significant, as apparent both from Table \ref{table:clarperformance} and previous work by \citet{mulcaire2018polyglot}. We observe that the arguments in CS are labeled at a significant finer granularity than those of other languages. 
For example, for temporal arguments alone, the argument set in Czech contains 9 different labels at the finest granularity. In contrast, each of the other languages has only one single label for temporal arguments. 
Since CLAR performs one-to-one mapping to and from the source language, we suspect that CLAR encounters challenges in choosing one among many fine grained arguments to map to a coarse argument in English. While it is possible to extend CLAR with many-to-one mapping, based on our preliminary study (Appendix \ref{ap:cs_excep}), 
it may introduce additional noise. We plan to explore this direction in the future.


\section{Related Work}
\label{sec:review}

Models for SRL largely fall into two categories: \textit{syntax-agnostic} and \textit{syntax-aware}. For a long time, syntax was considered a prerequisite for better SRL performance~\cite{punyakanok2008importance,gildea2002automatic}. In the absence of syntactic information, these methods struggle to capture the discriminatory features and thus perform poorly. 

Recently, end-to-end deep neural models have been shown to extract useful discriminatory features even without syntactic information \cite{zhou2015end,marcheggiani2017simple,tan2018deep,he2018jointly} and achieve state-of-the-art performance. However, some works \cite{roth2016neural,he2017deep,strubell2018linguistically} argue that given a high-quality syntax parser, it is possible to further improve the SRL performance. Along this line, \cite{marcheggiani2017encoding} proposed a SRL model based on graph convolutional networks which incorporates syntactic information from a parser \cite{kiperwasser2016simple}. Further, \cite{li2018unified} proposes a more general framework to integrate syntax into SRL tasks. All these methods have been shown to perform well on rich resource languages.

Several recent attempts have been made to transfer knowledge from rich source languages to low resource languages for SRL tasks \cite{mulcaire2018polyglot,mulcaire2019polyglot} such that the knowledge transfer helps the model to learn better feature representations for low resource languages. To some extent, in other NLP tasks such as named identity recognition \cite{xie2018neural}, and syntactic dependency parsing \cite{ammar2016many} this knowledge transfer seems to be helping low resource languages. Our experimental results further strengthen this claim and confirm that languages share knowledge at the semantic level as well. 

An alternative line of work transfers cross-lingual knowledge to generate semantic labels for low resource languages by exploiting the monolingual SRL model and Multilingual parallel data \cite{akbik2016towards,akbik2016polyglot} with an assumption that the sentences in parallel corpora are semantically equivalent. Similarly, \cite{prazak2017cross} converts the monolingual dependency tree to a universal dependency tree for cross-lingual transfer. Though these methods do not require the knowledge of semantic roles in the target language, they require the availability of massive parallel corpora. On the other hand, CLAR is able to detect the similarity among arguments between the language pairs even in the presence of less data.

\section{Conclusion}
\label{sec:conc}
We introduces CLAR, a Cross-Lingual Argument Regularizer. It explores linguistic annotation similarity across languages and exploits this obtained information during SRL model training to map the target language arguments as the deformation of a space on which source language arguments lie. We confirm the effectiveness of CLAR for SRL on CoNLL 2009 dataset over monolingual and polyglot methods, without prior knowledge of cross-lingual alignments or parallel data. This paper demonstrates the promise of understanding and exploiting linguistic annotation similarity across languages during polyglot training. We plan to explore other ways of identifying and leveraging linguistic annotation similarity across languages.

\section*{Acknowledgments}

The authors would like to thank Ranit Aharonov, Mo Yu, and Tyler Baldwin for their comments on an early draft of this work. We also thank our anonymous reviewers for their constructive comments and feedback.

\bibliography{anthology,acl2019}
\bibliographystyle{acl_natbib}

\appendix

\section{Dataset Description}
\label{ap:datadesc}

Table~\ref{table:datastat} describes the training data statistics for each language. In the dataset, for every language, all sentences are marked with predicate-argument structures.  Across the languages the argument label set is different. 

\begin{table*}[t!]
\footnotesize
\begin{center}
\begin{tabu}{lrcrccrrl}
\toprule
Dataset &Word 	&POS 	&Lemma		&Arg Labels &Pred Labels 	& ${\#}$ Predicate 	&${\#}$ Arguments & train/valid/test/ood	\\ \midrule
CA 		&31,079	&15		&22,388		&39			&14				& 37,431			&84,367 	&13K/1.7K/1.8K/-		\\ 
CS 		&75,572	&15		&35,310		&62			&116			& 414,237			&365,255 &38K/5.2K/4.2K/1.1K			\\ 
DE 		&67,548	&57		&48,217		&10			&28				& 17,400			&34,276 	& 36K/1.6K/1.7K/707		\\ 
EN 		&30,479	&49		&23,727		&53			&21				& 179,014			&393,699 &39K/1.3K/2.4K/425			\\ 
ES 		&37,908	&15		&24,157		&43			&13				& 43,824			&99,054		&14K/1.6K/1.7K/-		\\ 
ZH 		&40,351	&38		&40,351		&37			&10				& 102,813			&231,869 & 22K/1.7K/2.5K/-			\\ 

\bottomrule
\end{tabu}
\end{center}
\caption{Train data statistics for each language. Languages are coded with ISO 639-1 codes.}
\label{table:datastat}
\end{table*}

\section{Hyperparameters}
\label{ap:hyper}
In our experiments, we randomly initialize the word and lemma embedding of dimension $100$ each, the pos embedding of dimension $32$, and the flag embedding of dimension $16$. We use the same model parameters as mentioned in \cite{li2018unified}: a 4-layer BiLSTM with 512 dimensional hidden units and $0.1$ dropout rate for the sentence encoder. Our role labeler has 5 MLP highway layers with ReLU activations. We train the model with Adam optimizer \cite{kingma2014adam} and minimize the final categorical cross-entropy objective. We train each model for 20 epochs and use early stopping with patience 5 on target language development set. For all the experiments, we repeat with 3 different initialization and report the average F1 score along with precision and recall. 

\begin{table*}[t!]
\begin{center}
\begin{tabu}{llllll}
\toprule
\multicolumn{2}{c}{\multirow{1}{*}{EN-ES}}&\multicolumn{2}{c}{\multirow{1}{*}{EN-ZH}}&\multicolumn{2}{c}{\multirow{1}{*}{EN-DE}}\\ \midrule
ES&EN&ZH&EN&DE&EN\\ \cmidrule(lr){1-2} \cmidrule(lr){3-4} \cmidrule(lr){5-6}  
{\small{\texttt{ES-argM-adv}}} & {\small{\texttt{AM-ADV}}}&   {\small{\texttt{ZH-DIS}}}&{\small{\texttt{AM-DIS}}}&    {\small{\texttt{DE-A0}}}&  {\small{\texttt{A0}}} \\ 
{\small{\texttt{ES-argM-tmp}}} & {\small{\texttt{AM-TMP}}}&   {\small{\texttt{ZH-LOC}}}&  {\small{\texttt{AM-LOC}}}&    {\small{\texttt{DE-A4}}} & {\small{\texttt{AM-TMP }}}\\ 
{\small{\texttt{ES-argM-fin}}} & {\small{\texttt{AM-PNC}}}&  {\small{\texttt{ZH-C-A0}}}& {\small{\texttt{AM-REC}}}&    {\small{\texttt{DE-A1}}}&  {\small{\texttt{A1}}} \\
{\small{\texttt{ES-argM-cau}}} & {\small{\texttt{AM-CAU}}}&   {\small{\texttt{ZH-ADV}}}&  {\small{\texttt{AM-ADV}}}&    &   \\ 
{\small{\texttt{ES-argL-null}}}& {\small{\texttt{AM-REC}}}&   {\small{\texttt{ZH-A0}}}&   {\small{\texttt{A0}}}&      &   \\ 
{\small{\texttt{ES-arg2-ext}}} & {\small{\texttt{C-AM-DIR}}}&   {\small{\texttt{ZH-TMP}}}&  {\small{\texttt{AM-TMP}}}&    &   \\ 
{\small{\texttt{ES-arg0-agt}}} & {\small{\texttt{A0}}}&     {\small{\texttt{ZH-MNR}}}&  {\small{\texttt{AM-MNR}}}&    &  \\ 
{\small{\texttt{ES-arg1-pat}}} & {\small{\texttt{A1}}}&     {\small{\texttt{ZH-A2}}}&   {\small{\texttt{A2 }}}&       &   \\
{\small{\texttt{ES-argM-mnr}}} & {\small{\texttt{AM-MNR}}}&   {\small{\texttt{ZH-A1}}}&   {\small{\texttt{A1}}}&      &   \\
{\small{\texttt{ES-argM-loc}}} & {\small{\texttt{AM-LOC}}}&      &     &      &  \\ \bottomrule
\end{tabu}
\end{center}
\vspace{-0.1in}
\caption{Paired arguments in the source and the target language detected by pair matching algorithm during CLAR training.}
\label{table:all_args}
\end{table*}
\section{Paired Arguments}
\label{ap:pairedargs}

We present the list of matched arguments for source-target language pairs in Table \ref{table:all_args}. We observe that almost all the argument pairs have similar meaning: some are syntactically visible (e.g. {\small{\texttt{ES-argM-adv}}} in ES and {\small{\texttt{AM-ADV}}} in EN), whereas others are semantically similar (e.g. {\small{\texttt{ES-argM-fin}}} and {\small{\texttt{AM-PNC}}} having the same meaning {\small{\texttt{purpose}}}). 

\begin{figure}[t!]
 \centerline{\begin{minipage}{0.5\columnwidth}
	\begin{center}
	\includegraphics[width=0.6\linewidth]{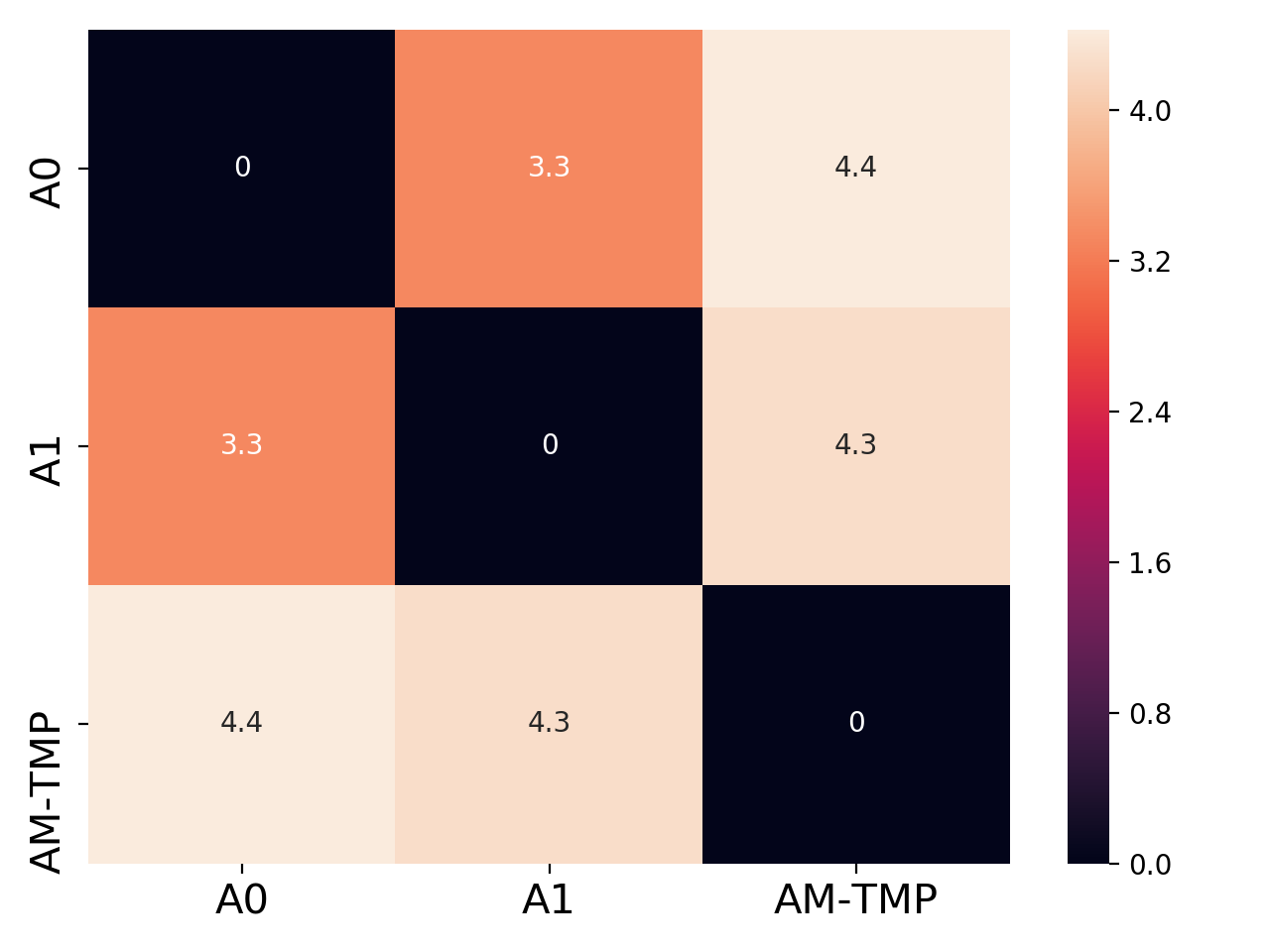}
	\end{center}
	\subcaption{EN}
	\label{fig::EN1}
\end{minipage}%
  \begin{minipage}{0.5\columnwidth}
	\begin{center}
	\includegraphics[width=0.6\linewidth]{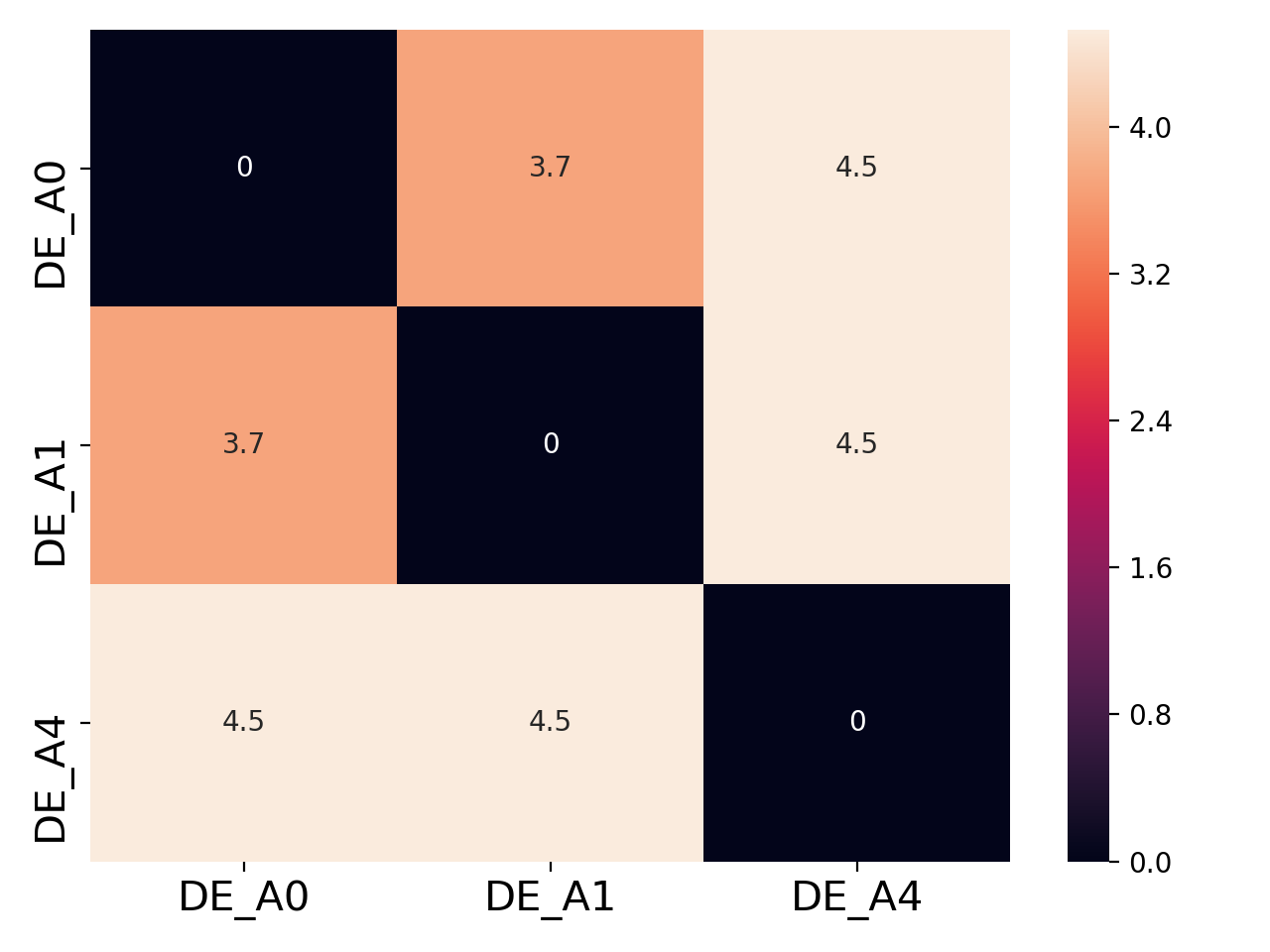}
	\end{center}
	\subcaption{DE}
	\label{fig::DE}
\end{minipage}}
~
  \begin{minipage}{0.5\columnwidth}
	\centering
	\includegraphics[width=1.1\linewidth]{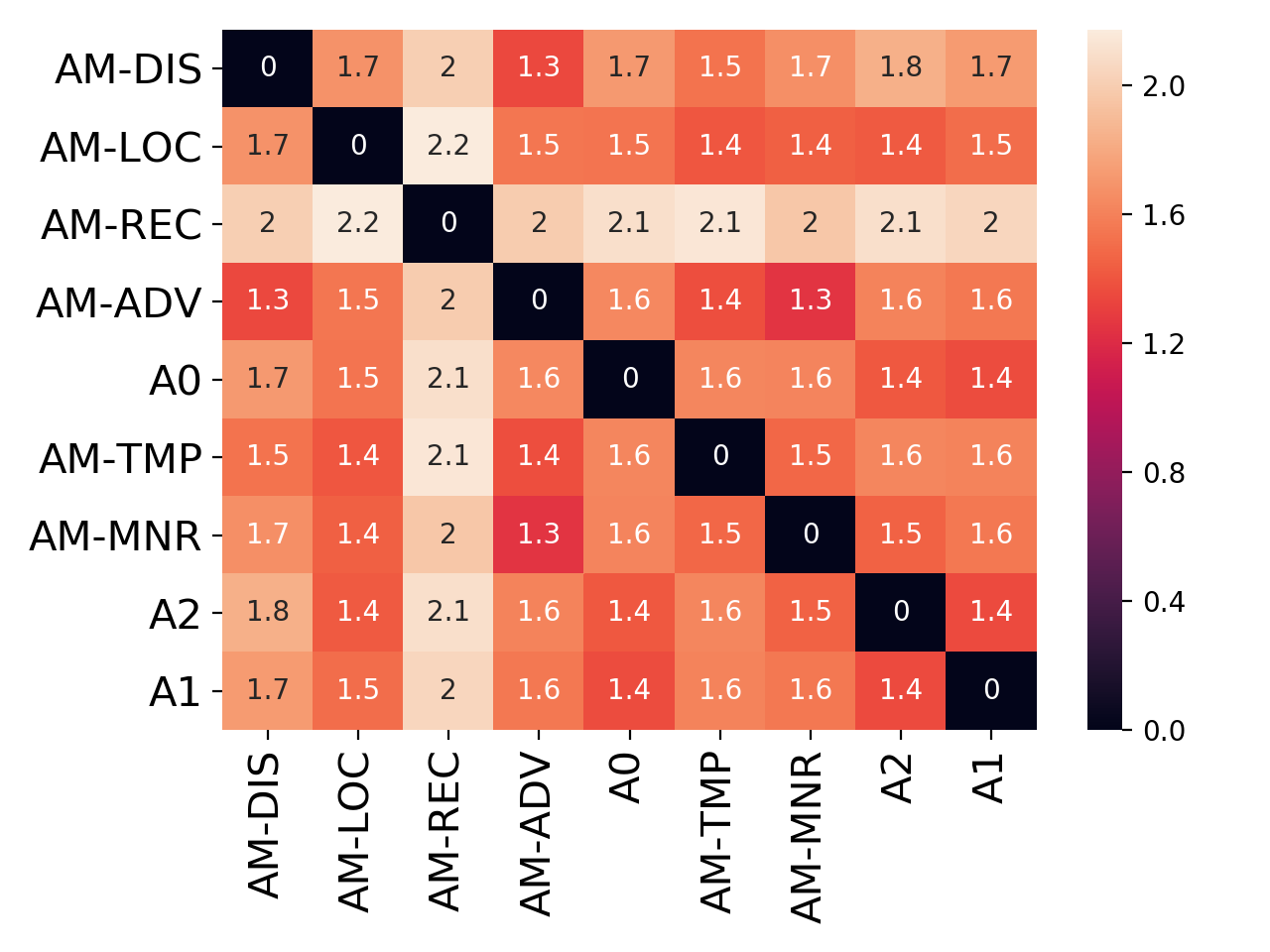}
	\subcaption{EN}
	\label{fig::EN2}
\end{minipage}%
  \begin{minipage}{0.5\columnwidth}
	\centering
	\includegraphics[width=1.1\linewidth]{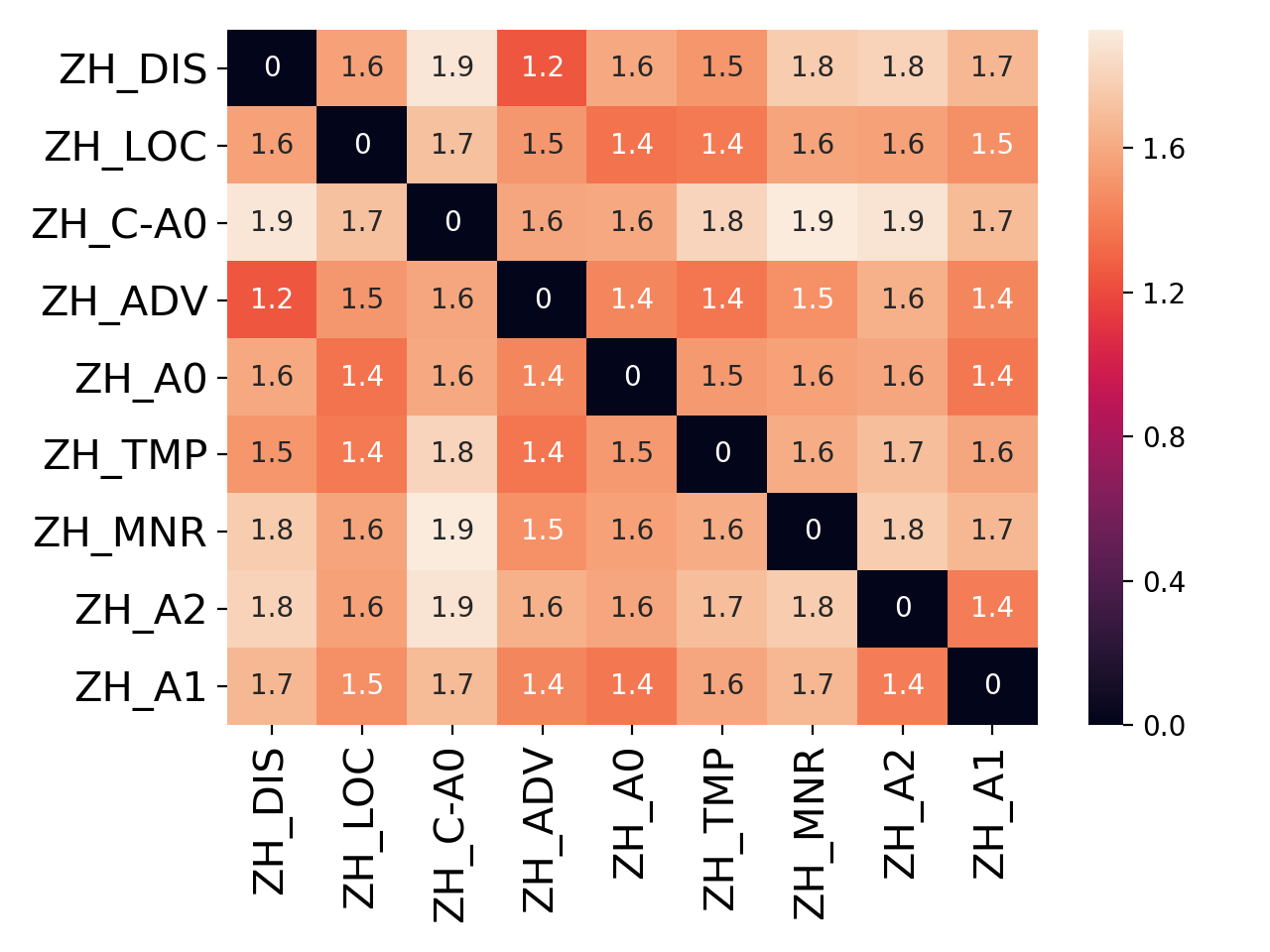}
	\subcaption{ZH}
	\label{fig::ZH}
\end{minipage}
~
  \begin{minipage}{0.5\columnwidth}
	\centering
	\includegraphics[width=1.1\linewidth]{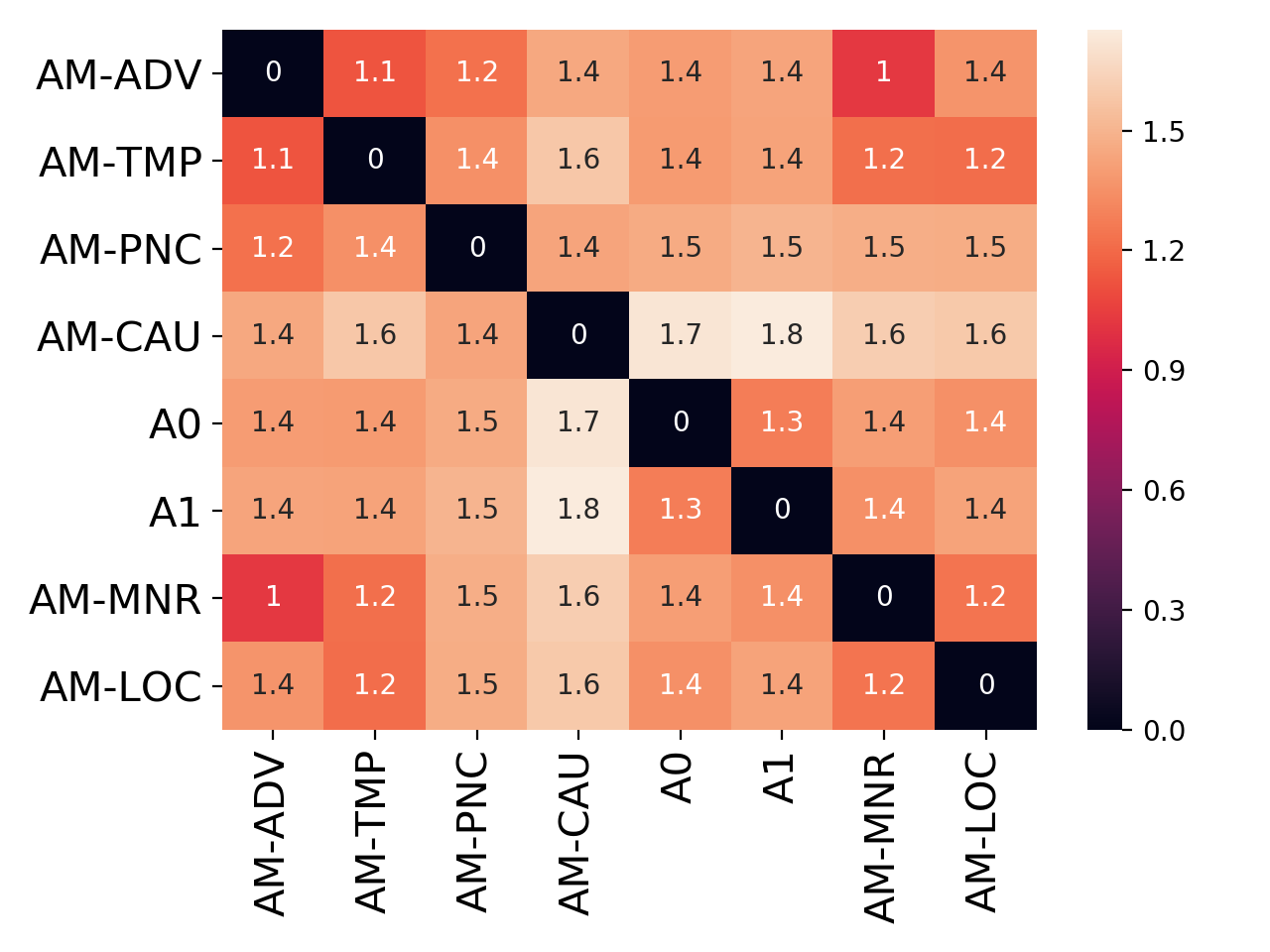}
	\subcaption{EN}
	\label{fig::EN3}
\end{minipage}%
  \begin{minipage}{0.5\columnwidth}
	\centering
	\includegraphics[width=1.1\linewidth]{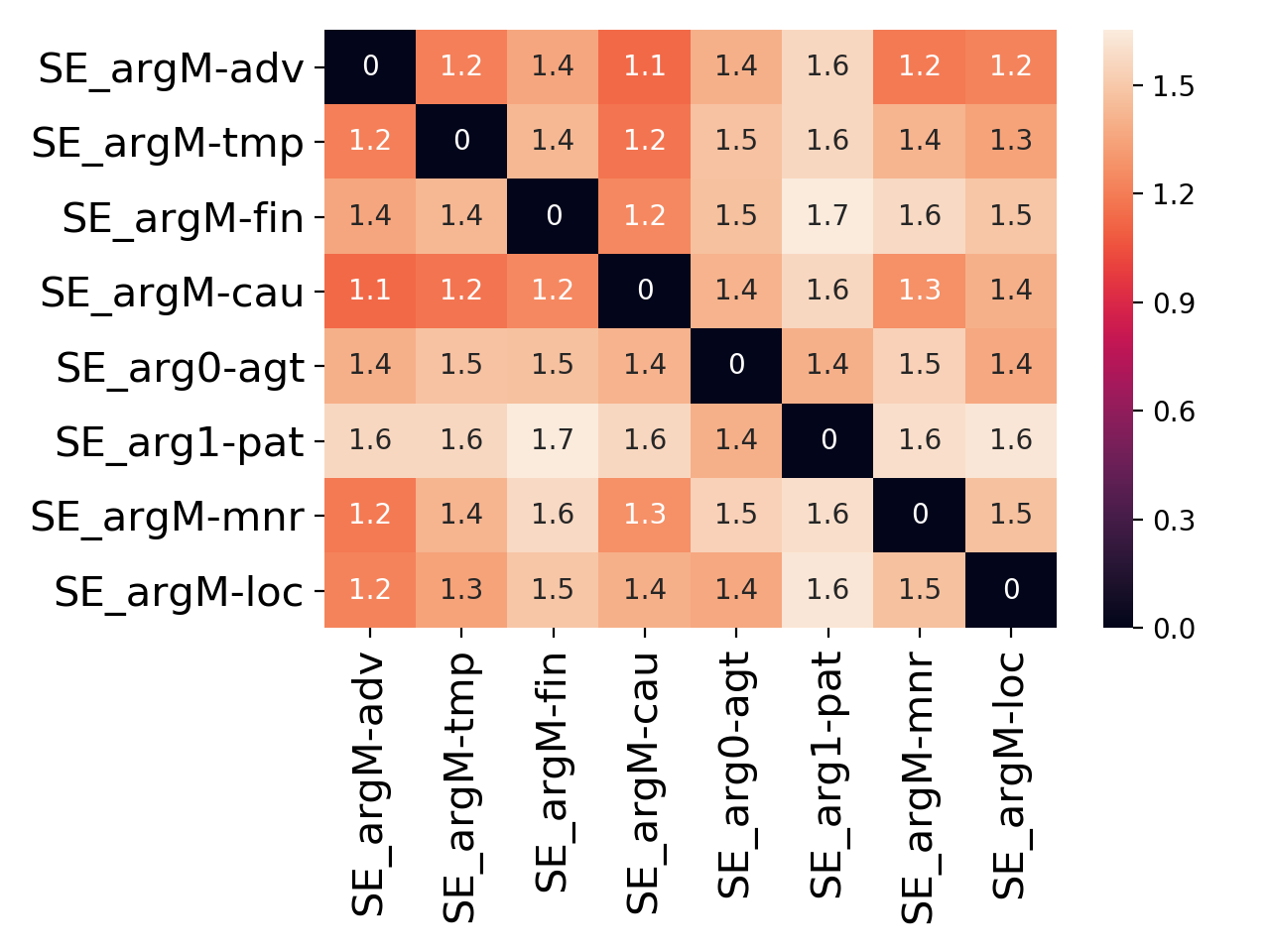}
	\subcaption{ES}
	\label{fig::ES}
\end{minipage}
\caption{Euclidean distance between last layer weights for matched arguments. Row I: EN-DE, Row II: EN-ZH, Row III: EN-ES, column I: Source language, column II: Target language}
\label{fig:distance}
\end{figure}

We also plot the the euclidean distance matrix among the matched source language arguments and among the target language arguments. In Fig.~\ref{fig:distance} we show the distance matrix for various language pairs. We compute the correlation coefficient between these matrices and All these coefficients are close to 1 which show that CLAR is indeed able to detect a manifold for the target language arguments similar to the one for the source language arguments.

\section{CLAR Extension to Many-to-one Mapping}
\label{ap:cs_excep}

We suspect that CLAR gets a difficulty in choosing one among many fine grained arguments to map to a coarse argument in source language. Here we perform the preliminary investigation on the many to one extension of CLAR.
Since CS have fine grained labels and is good candidate to analyze many to one mapping, we allow many-to-one argument mapping from Czech to English by relaxing a constraint in the final optimization function and updating  only this constraint
\begin{equation}
\sum_j T_{ij} \leq M,  \; i = 1, \ldots, \hat{k}_s,
\end{equation}
while keeping all the other constraints intact. This modification allows at most $M$ arguments in CS to pair with only one argument in EN. Now, following the training procedure, we observe that CLAR is able to efficiently capture many-to-one mappings with minimum noise.
\begin{table}[t!]
\footnotesize
\begin{center}
\begin{tabu}{lrlr}\toprule
CS &	EN &	CS &	EN\\  \cmidrule(lr){1-2} \cmidrule(lr){3-4}
{\small{\texttt{PAT}}}	&{\small{\texttt{A1}}} 	&{\small{\texttt{MAT}}}&{\small{\texttt{A3}}}\\ 
{\small{\texttt{ACT}}}	&{\small{\texttt{A0}}} 	&{\small{\texttt{BEN}}}&{\small{\texttt{A3}}}\\ 
{\small{\texttt{APP}}}	&{\small{\texttt{A2}}} 	&{\small{\texttt{ACMP}}}& {\small{\texttt{AM-ADV}}}\\ 
{\small{\texttt{ADDR}}}	&{\small{\texttt{A2}}} 	&{\small{\texttt{CAUS}}}& {\small{\texttt{AM-ADV}}}\\ 
{\small{\texttt{DIR3}}}	&{\small{\texttt{A2}}} 	&{\small{\texttt{COND}}}& {\small{\texttt{AM-ADV}}}\\ 
{\small{\texttt{TWHEN}}}& {\small{\texttt{AM-TMP}}}&{\small{\texttt{COMPL}}}& {\small{\texttt{AM-DIS}}}\\ 
{\small{\texttt{THL}}}	& {\small{\texttt{AM-TMP}}}&{\small{\texttt{CPHR}}}&{\small{\texttt{C-A1}}}\\ 
{\small{\texttt{THO}}}	& {\small{\texttt{AM-TMP}}}&{\small{\texttt{EFF}}}& {\small{\texttt{AM-PNC}}}\\ 
{\small{\texttt{MANN}}}& {\small{\texttt{AM-MNR}}}&{\small{\texttt{AIM}}}& {\small{\texttt{AM-PNC}}}\\ 
{\small{\texttt{REG}}}	& {\small{\texttt{AM-MNR}}}&{\small{\texttt{EXT}}}& {\small{\texttt{AM-EXT}}}\\ 
{\small{\texttt{MEANS}}}& {\small{\texttt{AM-MNR}}}&{\small{\texttt{DPHR}}}& {\small{\texttt{AM-DIR}}}\\ 
{\small{\texttt{LOC}}}	& {\small{\texttt{AM-LOC}}}&{\small{\texttt{CRIT}}}& {\small{\texttt{R-AM-TMP}}}\\ 
{\small{\texttt{RSTR}}}	& {\small{\texttt{AM-LOC}}}&{\small{\texttt{TTILL}}}& {\small{\texttt{R-AM-TMP}}}\\ 
{\small{\texttt{ID}}}	& {\small{\texttt{AM-LOC}}}&{\small{\texttt{TSIN}}}& {\small{\texttt{R-AM-TMP}}}\\
{\small{\texttt{COMPL2}}}& {\small{\texttt{AM-LOC}}}& & \\
{\small{\texttt{ORIG}}}	 & {\small{\texttt{AM-LOC}}}&  & \\ \midrule
\end{tabu}
\end{center}
\vspace{-0.1in}
\caption{Paired arguments in the source (EN) and the target language (CS)}
\label{table:ENCS_args}
\end{table}
In Table \ref{table:ENCS_args}, we present the argument pairs matched by CLAR. Interestingly, CLAR detects most of the argument pairs correctly, for example, \{{\small{\texttt{TWHEN}}}, {\small{\texttt{THL}}}, {\small{\texttt{THO}}}\} in CS are mapped to {\small{\texttt{AM-TMP}}} in EN, as expected. However, there are a few pairs that are wrongly mapped, for instance, {\small{\texttt{DIR3}}} in CS is mapped to {\small{\texttt{A2}}} in EN. We find that the detection of these noisy pairs is difficult to avoid as the \emph{Prague Dependency Treebank 2.0.} \cite{hajic2003pdt} (source of CS dataset) itself points the borderline cases associated with each argument label in CS. For example, {\small{\texttt{ACMP}}} in CS has borderline cases with both {\small{\texttt{COND}}} and {\small{\texttt{CAUS}}}, therefore, they are mapped together to a single argument in EN.  
\begin{table}[t!]
\footnotesize
\begin{center}
\begin{tabu}{lccc}\toprule
 \multicolumn{1}{l}{\multirow{1}{*}{CLAR Mapping}} & \textbf{P} & \textbf{R} & \textbf{F1}\\ \midrule
one-one 	&  79.91 &  {76.50}   &  {78.17}\\ 
many-one &  79.72  &  76.05  &  77.84\\ 
many-one (combined) &  82.57  &  75.40  &  \textbf{78.82}\\\midrule
\end{tabu}
\end{center}
\vspace{-0.15in}
\caption{Czech argument classification performance with many to one argument mapping.}
\label{table:ENCS}
\end{table}

Although CLAR with many-to-one mapping is able to match multiple target language argument labels to a single source language argument label, it actually leads to performance drop as compared to one-to-one mapping (Table \ref{table:ENCS}). This drop in performance is likely because while learning many-to-one mappings, CLAR loses its discriminatory power among those multiple arguments which are mapped to a single label. To validate this phenomenon, at test time, we combine all the argument labels mapped to a single label both for the target and the prediction set; that is, we combine {\{{\small{\texttt{TWHEN}}}, {\small{\texttt{THL}}}, {\small{\texttt{THO}}}\}} and propose a new label (say {\small{\texttt{TWHEN}}}) and observe 1ppt $\uparrow$ in $F_1$ on these combined labels. However, how to effectively leverage CLAR with many-to-one mapping for SRL model training remains an open question and requires further exploration in the future.


\end{document}